\documentclass[10pt,twocolumn,letterpaper]{article}

\usepackage{cvpr}              

\usepackage{multirow}
\usepackage{xcolor}
\usepackage{colortbl}
\usepackage{adjustbox}
\usepackage{makecell}
\usepackage{circledsteps}
\usepackage{amsthm} 
\pgfkeys{/csteps/black-circle/.style={
  inner color=white,
  outer color=black,
  fill color=black,
}}
\usepackage{cancel}
\usepackage{tikz}
\usepackage{pgfplots}\pgfplotsset{compat=1.9}

\newcommand{\name}{Logits Tempering Unlearning Strategy}

\newcommand{\std}[1]{\textsubscript{\scriptsize{$\pm$#1}}}
\newcommand{\bl}[1]{{\color{blue}(#1)}}
\newcommand{\B}[1]{{\color{blue}#1}}
\newcommand{\ul}[1]{{\underline{#1}}}
\newcommand{\tb}[1]{{\textbf{#1}}}
\newcommand{\smallcircle}[1]{\scalebox{0.8}{\Circled{#1}}}

\definecolor{cvprblue}{rgb}{0.21,0.49,0.74}
\usepackage[pagebackref,breaklinks,colorlinks,allcolors=cvprblue]{hyperref}

\def\paperID{16551} 
\def\confName{CVPR}
\def\confYear{2025}

\title{LoTUS: Large-Scale Machine Unlearning with a Taste of Uncertainty}

\author{
Christoforos N. Spartalis\textsuperscript{1,2}\hspace{1.2em}
Theodoros Semertzidis\textsuperscript{2}\hspace{1.2em}
Petros Daras\textsuperscript{2}\hspace{1.2em}
Efstratios Gavves\textsuperscript{1,3}\\
\textsuperscript{1}University of Amsterdam,
\textsuperscript{2}Centre for Research \& Technology Hellas,
\textsuperscript{3}Archimedes/Athena RC\\
{\tt\small \{c.spartalis,e.gavves\}@uva.nl} \hspace{1.2em}
{\tt\small \{c.spartalis,theosem,daras\}@iti.gr}
}

\begin{document}
\maketitle
\begin{abstract}
We present LoTUS,
a novel Machine Unlearning (MU) method that eliminates the influence of training samples from pre-trained models, avoiding retraining from scratch. LoTUS smooths the prediction probabilities of the model up to an information-theoretic bound, mitigating its over-confidence stemming from data memorization. We evaluate LoTUS on Transformer and ResNet18 models against eight baselines across five public datasets. Beyond established MU benchmarks, we evaluate unlearning on ImageNet1k, a large-scale dataset, where retraining is impractical, simulating real-world conditions. Moreover, we introduce the novel Retrain-Free Jensen-Shannon Divergence (RF-JSD) metric to enable evaluation under real-world conditions. The experimental results show that LoTUS outperforms state-of-the-art methods in terms of both efficiency and effectiveness. Code: \href{https://github.com/cspartalis/LoTUS}{https://github.com/cspartalis/LoTUS}.
\small{Conference on Computer Vision and Pattern Recognition, 2025.}
\looseness=-1
\end{abstract}    
\vspace{-1em}
\section{Introduction}
\label{sec:intro}

Machine Unlearning focuses on removing the influence of training samples from pre-trained models without retraining the model entirely \cite{liu2024machine}.
Its applications include privacy protection in Machine Learning~\cite{foster2024fast, golatkar2020eternal, bourtoule2021machine}.
%
%
As an alternative to retraining a model from scratch, Machine Unlearning addresses three principal challenges:
\smallcircle{1} minimizing the time window during which the model is vulnerable,
\smallcircle{2} minimizing the cost in terms of time and computational resources, and
\smallcircle{3} minimizing the dependency on access to all training data to retain the utility of the pre-trained model, as full data access is often limited due to privacy policies and storage limitations.
\looseness=-1
Therefore, an effective and efficient unlearning algorithm should meet the following requirements~\cite{foster2024information}:
\smallcircle{1} Effectively eliminate the impact of specific training samples from the model.
\smallcircle{2} Retain the model's performance on the remaining training samples, even if access to the training set is limited.
\smallcircle{3} Be efficient in terms of both time and computational resources.
\looseness=-1
%

Considering only the effectiveness of unlearning, the \textit{gold standard} is to retrain the model from scratch without the samples designated for unlearning (also known as forget samples). To this end, two main taxonomy classes have been developed: \textit{exact unlearning}, which aims to produce a model that is statistically indistinguishable from the gold standard, which is often infeasible for complex algorithms \cite{cao2015towards} or inefficient \cite{bourtoule2021machine}, and \textit{approximate unlearning}, which relaxes the constraints of exact unlearning and adopts a suite of evaluation metrics that typically measure how well the unlearned model approximates the gold standard in terms of accuracy and resilience to privacy attacks \cite{fan2024salun}.
The scope of this study concerns the following questions:

\vspace{-0.45em}
\begin{quote}
\textbf{Q1:} Can an unlearning method efficiently eliminate the influence of training samples from a pre-trained model while approximating the effectiveness of the gold standard?
\looseness=-1
\end{quote}
\vspace{-1em}
\begin{quote}
\textbf{Q2:} Can this unlearning method effectively handle large-scale datasets and models under real-world constraints, including limited data access?
\looseness=-1
\end{quote}
\vspace{-0.45em}

\begin{figure}[t]
  \centering
   \includegraphics[width=1\columnwidth]{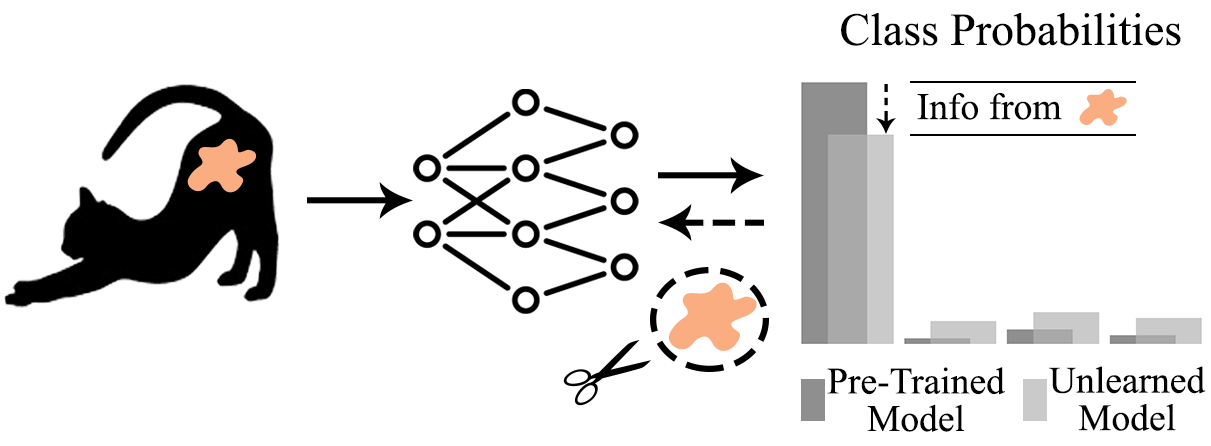}
   \caption{\textbf{Machine Unlearning via smoothing prediction probabilities:} LoTUS eliminates sample-specific information (e.g., unique fur patterns in cat images) that the DNN memorized and exposed through overconfident predictions. Then, the DNN responds to unlearned samples as if they were never part of the training set.}
   \label{fig:fig1}
\end{figure}

To answer these questions we propose the \ul{Lo}gits \ul{T}empering \ul{U}nlearning \ul{S}trategy (\textit{LoTUS} for short, such as the fruit that made Ulysses' comrades forget).
LoTUS leverages the known tendency of Deep Neural Networks (DNNs) to memorize sample-specific features from training data and output over-confident predictions~\cite{ye2024leave}, a vulnerability exploited by Membership Inference Attacks (MIAs) to assess whether a sample is a member of the training set~\cite{shokri2021privacy}.
To this end, LoTUS smooths the model's output probabilities, as shown in \cref{fig:fig1}, increasing the entropy to resemble that of unseen (during training) samples.
This unseen set, which may include synthetic data, enables LoTUS to calibrate the retained information for forget samples post-unlearning and replicate the decision-making process of the gold standard model. 
Since the gold standard model was not trained on the forget samples, it naturally avoids overconfident predictions and typically exhibits lower accuracy on them.
To better approximate the gold standard's performance, LoTUS also introduces Gumbel noise into the pre-trained model's output distribution. This encourages diverse predictions and helps reduce the pre-trained model's accuracy on forget samples to resemble that of the gold standard.
\looseness=-1

%


In contrast to previous studies that have focused mainly on the input or model space, LoTUS follows an \textit{entropy-based} approach that directly modifies the model's output probabilities, emphasizing an underexplored unlearning approach.  
The difference from the existing method which indiscriminately maximizes entropy using random labeling \cite{graves2021amnesiac} is that LoTUS uses an \textit{information-theoretical} bound to control the uncertainty introduced to the model.
\looseness=-1
Our main contributions are as follows:
\begin{enumerate}
\setlength{\itemsep}{0.5em}
\item{We introduce LoTUS, the first unlearning method that operates directly in the model's output space, while following an information-theoretic approach to determine the amount of entropy increase. This bound enables cautious unlearning that approximates the gold standard.\looseness=-1}
%
%
\item{We introduce the \textit{Retrain-Free Jensen-Shannon Divergence (RF-JSD)} unlearning metric, which enables evaluation in real-world scenarios.
RF-JSD exhibits a strong Pearson correlation ($\text{PCC}\!=\!0.92\std{0.04}$) with the established JSD score while eliminating the need to retrain the model.
Compared to the existing retrain-free ZRF score, RF-JSD offers enhanced interpretability and efficiency.\looseness=-1}
%
\item{
We introduce a novel large-scale experimental setup that incorporates a large-scale dataset (ImageNet1k), and limited access to the training set, with the aim of simulating real-world conditions, where model retraining is infeasible.
%
%
%
%
Overall, we evaluated LoTUS on the Vision Tranformer and ResNet18 modes, against eight baseline methods on five public datasets.
Extensive experiments demonstrate that LoTUS outperforms state-of-the-art approaches in terms of both unlearning effectiveness and efficiency, in all benchmarks (novel and established).}
%
\looseness=-1
\end{enumerate}
%

\section{Related Work}

\paragraph{Machine Unlearning}
was first introduced in~\cite{cao2015towards} with an approach that decomposes traditional Machine Learning algorithms into summations, enabling the reduction of the influence of specific data points for exact unlearning. Subsequently, a theoretical framework for approximate unlearning was proposed in~\cite{guo2020certified}, suggesting a Hessian-based regularization technique limited to models with convex loss functions to mitigate membership inference risks. Unlearning was subsequently extended to deep neural networks in~\cite{golatkar2020eternal} by introducing a Lagrangian regularization approach that utilizes the Fisher Information Matrix as a Hessian approximation. More recent works have improved unlearning effectiveness and efficiency~\cite{kurmanji2024towards, foster2024fast} and expanded machine unlearning applications in diverse areas, including user privacy~\cite{seo2024generative}, security defense~\cite{schoepf2024potion}, toxic content removal~\cite{fan2024salun, heng2024selective}, copyright protection~\cite{golatkar2024cpr}, and bias mitigation~\cite{chen2024fast}. Emphasizing 
 on privacy applications, Machine Unlearning has been defined as a privacy game aiming to reduce the accuracy of MIAs~\cite{bourtoule2021machine}. 
 \looseness=-1

Algorithms are categorized into two classes depending on where the manipulations are applied: model space and data space~\cite{xu2024machine}. In model-space approaches, manipulations include regularizing the loss function to shift model weights far from the pre-trained model and close to the gold standard~\cite{kurmanji2024towards, chundawat2023can, golatkar2020eternal}. Another approach is pruning, which involves identifying and reducing the influence of weights that are most affected by the data to be unlearned~\cite{foster2024fast, fan2024salun}. Although model-space approaches can offer theoretical justifications and efficiency, they also present challenges in terms of implementation complexity and interpretability~\cite{xu2024machine}. 
\looseness=-1

On the other hand, data-space approaches focus on reorganizing or modifying the data to be unlearned. These methods include data-partitioning techniques that track which partition each data point belongs to and the corresponding model updates they trigger. They enable selective forgetting by isolating specific model updates~\cite{graves2021amnesiac, cao2015towards} or by retraining the model from the latest valid checkpoint~\cite{bourtoule2021machine}. These techniques are usually pre-hoc; meaning they must be applied before training, and cannot apply to pre-trained models. Also, they are resource-intensive, trading increased space complexity for reduced time complexity.
\looseness=-1

Data obfuscation is a data-space approach that can be applied post-hoc. This includes methods such as adversarial attacks~\cite{cha2024learning, chen2023boundary} or adding noise
to the input~\cite{foster2024information, chundawat2023can}. Although these techniques primarily focus on modifications in the input space, Random Labeling~\cite{graves2021amnesiac} takes a different approach by altering the output space and reassigning incorrect labels to the forget samples. Despite its simplicity, this approach has been shown to be effective~\cite{triantafillou2024we}. Data-space adjustments are more conceptually aligned with information theory, although a direct connection was explicitly established only recently in~\cite{foster2024information}, which explores input perturbations.
\looseness=-1
\vspace{-1.5em}
\paragraph{Information Theory} formalizes the quantification of information through mathematical measures such as entropy and mutual information~\cite{shannon1948mathematical}.
In the context of DNNs, information is typically defined for random variables such as the input and output of the models.


\section{\name}

\subsection{Preliminaries}
Let ${x\!\sim\!P(X)}$ be a feature vector representing an image sampled from the sampling space $A(X)$,
%
and ${y\!\sim\!P(Y)}$ be a classification label sampled from the sampling space ${A(Y)\!=\!\{c_1,c_2,\!\dots\!,c_k\}}$, where $k$ is the total number of classes.
Let ${f_w(X)\!: A(X) \!\mapsto\! A(Y)}$ be a DNN model parameterized by weights $w$ that maps an image $x$ to a classification label $y$.
%
%
Also, let ${D\!=\!\{(x_i,y_i)\}_{i=1}^n\!=\! D_f \!\cup\! D_r \!\cup\! D_u}$ be a dataset of images $x_i \!\in\! A(X)$ and their corresponding labels $y_i \!\in\! A(Y)$, which comprises three pairwise disjoint datasets:
\smallcircle{1}~\textbf{Forget set \textit{D\textsubscript{f}}}: Training samples whose influence on the model weights $w$ should be removed.
\smallcircle{2}~\textbf{Retain set \textit{D\textsubscript{r}}}: Training samples whose influence on $w$ must be preserved.
\smallcircle{3}~\textbf{Unseen set \textit{D\textsubscript{u}}}: Samples that were not used to train the model $f_w$.
As unseen sets, we use either the validation sets or synthetic data generated from training data.
Finally, we denote:
$f_\text{orig}$ as the pre-trained (or original) model, trained on $D_f \cup D_r$,
$f_\text{gold}$ as the gold standard model, retrained from scratch only on $D_r$, and
$f_\text{un}$ as the model derived from unlearning, which is the process of updating the model weights of $f_{\text{orig}}$ so that ${f_{\text{un}}(x) \!\approx\! f_{\text{gold}}(x), \, \forall x \!\in\! D}$.
\looseness=-1

\subsection{Upper-bounding Uncertainty}\label{sec:upper_bound}
Unlike existing unlearning methods~\cite{graves2021amnesiac}, which indiscriminately increase entropy in the output space, we aim to establish an upper bound on the uncertainty introduced by unlearning, removing only the information specific to the forget set $D_f$ which extends beyond the model's general knowledge.
%
%
To achieve this, we adopt an information-theoretic framework to delineate the information essential for preserving model utility from the information that needs to be removed. Although directly estimating the mutual information between the model's input and output would be ideal, this approach is both challenging and computationally intensive~\cite{belghazi2018mutual}. Therefore, we introduce a relaxed version of the framework that enables the assessment of the appropriate entropy increase required for unlearning.
\looseness=-1

\paragraph{Proposition.}
Let $X_s$ be a random variable with any sampling space ${A(X_s) \!\subset\! A(X)}$. In other words, $X_s$ is derived from $X$ by filtering.
Then, ${X \!\to\! X_s \!\to\! f_w(X_s)}$ is a processing chain where $f_w(X_s)$ depends on $X$ only through $X_s$.
By the Data Processing Inequality~\cite{thomas2012elements}, this is a Markov chain that implies $f_w(X_s) \!\to\! X_s \!\to\! X$. Therefore, by the chain rule, we can expand the mutual information in two different ways:
\looseness=-1
\begin{equation}
    \begin{split}
        I\big(f_w(X_s);X_s,X\big) \!&=\! I\big(f_w(X_s);X\big) + I\big(f_w(X_s);X_s\!\mid\!X\big) \\
        \!&=\! I\big(f_w(X_s); X_s) + \cancelto{0}{I\big(f_w(X_s); X \!\mid\! X_s\!\big)}
    \end{split}
    \label{eq:chain_rule}
\end{equation}
Since $f_w(X_s)$ is conditionally independent of $X$ given $X_s$, it follows that $I\big(f_w(X_s); X \!\mid\! X_s\!\big) \!=\! 0$. Therefore, from~\cref{eq:chain_rule}, the mutual information between the input $X_s$ and the output $f_w(X_s)$ of the classifier is:
\looseness=-1
\begin{equation}
	\underbrace{I(f_w(X_s); X_s)}_{\substack{\text{total information captured by} \\\text{the model from the data subset}}} \!=\! \underbrace{I(f_w(X_s); X)}_{\substack{\text{global}\\\text{information}}} + \underbrace{I(f_w(X_s); X_s \!\mid\! X)}_{\substack{\text{subset-specific}\\\text{information}}} 
 \label{eq:main}
\end{equation}

We consider $I(f_w(X_s); X)$ as the \textit{global information} a model $f_w$ has captured from the set $A(X)$. In other words, it quantifies the contribution of the shared features among training samples in $A(X)$ (\ie, global features) to the model's decision-making.
Respectively, we consider $I(f_w(X_s); X_s \!\mid\! X)$ as the additional \textit{subset-specific information} learned exclusively from the subset $A(X_s)$, which refines the model's decision and adds detail beyond what is already captured from $A(X)$.
\looseness=-1
 
For example, if there are images of \emph{cats} in both $A(X)$ and its subset $A(X_s)$, then the \emph{total information} captured from the images in $A(X_s)$ can be categorized into two types: The \emph{global information} learned from shared features across all cat images in $A(X)$, \eg body shape of cats; and the additional \emph{subset-specific information} learned exclusively from cat images in $A(X_s)$, \eg unique fur patterns. 
\looseness=-1

To determine the presence of \emph{subset-specific information} and how this is expressed in the model's decision, we refer to the memorization capabilities of DNNs and the derived privacy considerations.
DNNs are known to memorize information from individual samples in the training set~\cite{ye2024leave}. Considering a DNN classifier, the memorization of specific patterns is exposed in the model's output probabilities via increased confidence (\ie, lower entropy in the model's output probability distribution), and this is an indicator exploited by privacy attacks to distinguish which samples are members of the training set~\cite{shokri2017membership}.
\looseness=-1

Therefore, if $A(X_s)$ is a subset of the training set, then the model can capture the \emph{subset-specific information} leading to over-confident predictions. However, if $A(X_s)$ was unseen during training, then the model had no chance to capture \emph{subset-specific information} and its predictions are based solely on the \emph{global information} captured from training samples in $A(X)$. Defining the sampling space of $X$ as $A(X)\!=\!D\!=\!D_f\cup D_r\cup D_u$, and the sampling space of $X_s^f$ as the forget set $A(X_s^f)\!=\!D_f \!\subset\! D$, we can assess the \emph{total information} captured by the available pre-trained model $f_\text{orig}$ and the ideal gold standard model $f_\text{gold}$ as such:
\looseness=-1
\begin{equation}\label{eq:i_original}
   I\big(\!f_\text{orig}(\!X_s^f\!); X_s^f ) \!=\! I\big(\!f_\text{orig}(\!X_s^f\!); X\!\big) + I\big(\!f_\text{orig}(\!X_s^f\!); X_s^f \!\mid\! X\!\big)
\end{equation}
\begin{equation}\label{eq:i_gold}
   I\big(\!f_\text{gold}(\!X_s^f\!); X_s^f \!\big) \!=\! I\big(\!f_\text{gold}(\!X_s^f\!); X\big) + \cancelto{0}{I\big(\!f_\text{gold}(\!X_s^f\!); X_s^f \!\mid\! X\!\big)}
\end{equation}
%
%
The gold standard model $f_\text{gold}$ has not been trained on the forget set $D_f$, thus $f_\text{gold}$ has not captured \emph{subset-specific information} from $D_f$ as shown in~\cref{eq:i_gold}.

Based on~\cref{eq:i_original,eq:i_gold}, we define Machine Unlearning as the process of eliminating the \textit{subset-specific information} ${I\big(\!f_\text{orig}(\!X_s^f); \!X_s^f \!\mid\! \!X)}$ from the pre-trained model (\ie, forgetting objective),
while retaining the \textit{global information} ${I\big(\!f_\text{orig}(\!X_s^f); \!X \big)}$ captured from the training samples in $D$ (\ie, retention objective to preserve model's utility on the remaining training samples).
Therefore, the \emph{total information} that the unlearned model $f_\text{un}$ retains for the samples in the forget set $A(X_s^f)\!=\! D_f$ is by definition equal to the \emph{global information} the pre-trained model $f_\text{orig}$ had captured from the training set $A(X) = D_f\!\cup\!D_r$:
%
\looseness=-1
\begin{equation}
    I\big(f_\text{un}(X_s^f); X_s^f\big) \!\overset{\Delta}{=}\! I\big(f_\text{un}(X_s^f); X\big) \overset{\Delta}{=} I\big(f_\text{orig}(X_s^f); X\big)
    \label{eq:objective}
\end{equation}
\ul{Assumption of instance-wise unlearning}: \Cref{eq:objective} holds under the condition that the forget set $D_f$ comprises only a subset of the training samples of a class (\ie, instance-wise unlearning) and not all class samples (\ie, class unlearning). For example, if $D_f$ contains images of cats, then the retain set $D_r$ must also include images of cats to ensure that the global features related to the cat class are still encoded in the model after unlearning. Otherwise, the \emph{global information} will be eliminated during unlearning and $I(f_\text{orig}(X_s^f); X) \!\neq\! I\big(f_\text{un}(X_s^f); X\big) \!=\! 0$.
\looseness=-1

Subsequently, we focus on instance-wise unlearning and the quantification of the \emph{global information} that should be retained post-unlearning.
However, in \cref{sec:class}, we provide details on the class unlearning task and how LoTUS can be easily adapted to this.
\looseness=-1

\paragraph{Quantifying \textit{global information}.}
Estimating the \emph{global information} ${I\big(f_\text{orig}(X_s^f); X\big)}$ is challenging due to the high dimensionality and complex dependencies in data, making it difficult and computationally intensive~\cite{belghazi2018mutual}.
%
To address this, we use the \textit{total information} ${I\big(f_\text{orig}(X_s^u); X_s^u\big)}$, where $A(X_s^u)\!=\!D_u$, as a proxy of the \textit{global information} and conclude on an efficient yet effective weaker approximation.  
\looseness=-1

As previously explained and shown in \cref{eq:i_gold}, a model cannot capture \emph{subset-specific information}, if this subset has not been used during training. Therefore, if $D_u$ consists of unseen (during training) samples, then:
\begin{equation}\label{eq:i_origDu}
   I\big(\!f_\text{orig}(\!X_s^u\!); X_s^u \!\big) \!=\! I\big(\!f_\text{orig}(\!X_s^u\!); X\big) + \cancelto{0}{I\big(\!f_\text{orig}(\!X_s^u\!); X_s^u \!\mid\! X\!\big)}
\end{equation}
%
%
\ul{Assumption of Distributional Similarity:}
The samples in the forget set $D_f$ and unseen set $D_u$ are assumed to follow the same distribution in terms of visual features and class distribution. This leads to the conclusion that their entropies are equal: $H(X_s^f)\!=\!H(X_s^u)$. Additionally, the \emph{total information} captured by the unlearned model $f_\text{un}$ from $D_f$ can be considered equivalent to that captured by the pre-trained model $f_\text{orig}$ from $D_u$. Based on \cref{eq:objective,eq:i_origDu}, we can thus reformulate the unlearning objective as:
\looseness=-1
\begin{equation}
\begin{gathered}
   I\big(f_\text{un}(X_s^f); X_s^f\big)= I\big(f_\text{orig}(X_s^u); X_s^u\big) \Rightarrow \\
   \cancel{H(X_s^f)} \!-\! H\big(X_s^f\!\mid\!f_\text{un}(X_s^f)\big) \!=\!
   \cancel{H(X_s^u)} \!-\! H\big(X_s^u\!\mid\!f_\text{orig}(X_s^u)\big) \Rightarrow \\
   H\big(X_s^f\!\mid\!f_\text{un}(X_s^f)\big) \!=\! H\big(X_s^u\!\mid\!f_\text{orig}(X_s^u)\big)
   \label{eq:cond_entropies}
\end{gathered}
\end{equation}
which establishes that, given the respective model's output, the uncertainty in identifying a forget versus an unseen sample is equal.
\looseness=-1

Although this theoretical formulation assumes an identical distribution for samples in $D_f$ and $D_u$, we show that the assumption can be relaxed in practice. Specifically, the images in both sets only need to share relevant features that contribute to the \emph{global information}. In other words, the forget and unseen sets should contain visually similar images rather than images with exactly the same information. For example, if the forget set contains cat images, the unseen set should also contain cat images --even synthetic ones-- rather than entirely different objects such as human portraits. In practice, this ensures sufficient similarity in global features related to the cat class.
\looseness=-1

\paragraph{Approximating conditional entropy.} Given the complexity of the underlying distributions and the computational challenge associated with entropy estimation in \cref{eq:cond_entropies}, we derive a practical relationship linking the prediction error to the uncertainty in the model's predictions. 
Let $\widehat X_s$ be an estimate of $X_s$ based on the model's output $f_w(X_s)$, following $X_s \!\to\! f_w(X_s) \!\to\! \widehat{X}_s$, and define the prediction error probability as ${P_e \!=\! P\{X_s \!\neq\! \widehat X_s\}}$. Then Fano's Inequality~\cite{thomas2012elements} states:
\looseness=-1
\begin{equation}
    P_e \geq \frac{H\big(X_s \mid f_w(X_s)\big) - 1}{\log |A(X_s)|}
\end{equation}
This inequality implies that lower prediction error $P_e$ --or equivalently higher accuracy (${\text{Acc}\!=\!1\!-\!P_e}$)-- corresponds to reduced uncertainty $H\big(X_s \mid f_w(X_s)\big)$.
Moreover, this aligns with the empirical observation that models tend to be more accurate in images for which they make predictions of higher confidence~\cite{wang2020tent}.
 
Since accuracy is straightforward to measure and computationally efficient, we approximate the conditional entropies in~\cref{eq:cond_entropies} and define a relaxed unlearning objective:
\begin{equation}
\text{Acc}(f_\text{un}, D_f) = \text{Acc}(f_\text{orig}, D_u)
\label{eq:acc_objective}
\end{equation}
where ${\text{Acc}(f_\Box, D_\triangle)}$ denotes the prediction accuracy of a model $f_\Box$ on a data subset $D_\triangle$.  \Cref{eq:acc_objective} suggests that the unlearning process can be calibrated by aligning the accuracy of the unlearned model on the forget set with the accuracy of the pre-trained model on the unseen set.

\subsection{Unlearning with LoTUS}

LoTUS leverages the accuracy objective in \cref{eq:acc_objective} to regulate the increase in model's uncertainty. Specifically, LoTUS increases model's uncertainty by smoothing the predicted probabilities of forget samples to tackle memorization --evident in over-confident predictions-- ensuring that the accuracy of the unlearned model $f_\text{un}$ converges toward that of the pre-trained model $f_\text{orig}$ on the unseen (during training) set. This approach not only eliminates the \emph{subset-specific information}, but also preserves the \emph{global information}, preventing over-unlearning and safeguarding the utility of the model on the remaining training samples.
\looseness=-1

%


%
%
%
%

To achieve this, LoTUS employs a knowledge distillation framework in which both the teacher and student models are initialized with the weights of the original model $f_\text{orig}$, as in~\cite{kurmanji2024towards}. The teacher serves as the original model $f_\text{orig}$ throughout the unlearning process, while the student $f_\text{un}$ undergoes unlearning by receiving perturbed knowledge from the teacher. This perturbation is applied during the activation of the teacher's logits using the Gumbel-Softmax function $gs(\cdot)$:
\looseness=-1
\begin{equation}
	p_i = gs(\pi, \tau) = \frac{\exp \left( \left( \log \pi_i \!+\! g_i \right) / \tau \right)}{ \sum_{j=1}^k \exp \left( \left( \log \pi_j \!+\! g_j \right) / \tau \right)} ,  \; i = 1,\!\dots\!, k
 \label{eq:probabilities}
\end{equation}
where $p_i$ is the probability of class $i$, $\pi_i$ is the corresponding logit, $g_i$ is statistical noise sampled from the Gumbel distribution, $k$ is the total number of classes, and $\tau\!\in\!\mathbb{R}^+$ is a temperature parameter that controls the sharpness of the output probabilities: smoothing them when ${\tau \!>\! 1}$, sharpening them when ${\tau \!<\! 1}$, and leaving them unchanged when ${\tau\!=\!1}$.
\looseness=-1

Temperature $\tau$ is the key component in LoTUS, as it controls the uncertainty introduced to the student by adjusting the entropy in the teacher's output probabilities.
In each unlearning epoch, the temperature $\tau$ is dynamically adjusted based on~\cref{eq:acc_objective} as follows:
\looseness=-1
\begin{equation}
    \tau_d = \exp\big({ \alpha \cdot (\text{Acc}(f_\text{un}, D_f) - \text{Acc}(f_\text{orig}, D_u))}\big)
    \label{eq:tau_d}
\end{equation}
where $f_\text{un}$ and $f_\text{orig}$ are the student and teacher models, respectively; $D_f$ is the forget set and $D_u$ the unseen set (\ie, validation set or synthetic data); and $\alpha\!\in\!\mathbb{R}^+$ is a positive value that scales the accuracy difference.
\looseness=-1

\begin{figure}[]
\begin{tikzpicture}
    \begin{axis}[
        width=8cm,
        height=3cm,
        xlabel={epochs},
        ylabel={$\Delta$Acc},
        xmin=0.5, xmax=10.5,
        ymin=-0.05, ymax=0.45,
        xtick={1,2,3,4,5,6,7,8,9, 10},
        ytick={-0.1,0,0.1,0.2,0.3,0.4,0.5},
        ymajorgrids=true,
        grid style=dashed,
        ticklabel style={font=\fontsize{7}{7}\selectfont}, 
        label style={font=\fontsize{9}{9}\selectfont}, 
        title style={font=\fontsize{9}{9}\selectfont}, 
        xlabel shift=-1.5ex 
    ]

    \addplot[
        color=blue,
        mark=square,
    ]
    coordinates {
        (1,0.4)(2,0.2)(3,0.08)(4,-0.02)(5,0.35)(6,0.06)(7,-0.03)(8,0.32)(9,0.12)(10,0.03)
    };
    
    \node[anchor=west, font=\fontsize{8}{8}\selectfont, draw,circle,inner sep=0.5pt] at (axis cs:1.5,0.33) {1};
    \node[anchor=west, font=\fontsize{8}{8}\selectfont, draw,circle,inner sep=0.5pt] at (axis cs:2.5,0.18) {2};
    \node[anchor=west, font=\fontsize{8}{8}\selectfont, draw,circle,inner sep=0.5pt] at (axis cs:3.4,0.08) {2};
    \node[anchor=west, font=\fontsize{8}{8}\selectfont, draw,circle,inner sep=0.5pt] at (axis cs:4.5,0.15) {3};
    \node[anchor=west, font=\fontsize{8}{8}\selectfont, draw,circle,inner sep=0.5pt] at (axis cs:5.55,0.24) {1};
    \node[anchor=west, font=\fontsize{8}{8}\selectfont, draw,circle,inner sep=0.5pt] at (axis cs:6.4,0.06) {2};
    \node[anchor=west, font=\fontsize{8}{8}\selectfont, draw,circle,inner sep=0.5pt] at (axis cs:7.6,0.15) {3};
    \node[anchor=west, font=\fontsize{8}{8}\selectfont, draw,circle,inner sep=0.5pt] at (axis cs:8.6,0.24) {1};
    \node[anchor=west, font=\fontsize{8}{8}\selectfont, draw,circle,inner sep=0.5pt] at (axis cs:9.5,0.11) {2};
    \end{axis}
\end{tikzpicture}
\vspace{-1em}
\caption{Contribution of the dynamically adjusted temperature $\tau_d$ to convergence toward the objective $\Delta\text{Acc}\!=\text{Acc}(f_\text{un}, D_f)\!-\!\text{Acc}(f_\text{orig}, D_u)\!=\!0$. The steps are denoted as follows: \smallcircle{1}: sharp step towards objective, \smallcircle{2}: smaller step (proportional to $\Delta\text{Acc}$), \smallcircle{3}: drastic accuracy restoration when over-unlearning.\looseness=-1}
\label{fig:tau}
\vspace{-1.25em}
\end{figure}
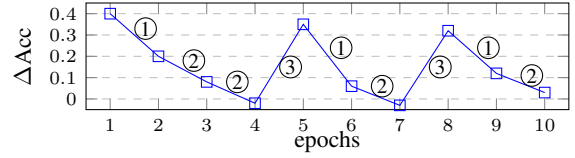

This implementation facilitates convergence to the unlearning objective~\cref{eq:acc_objective} by dynamically adjusting the entropy in the teacher's output probabilities as follows: \smallcircle{1} At the beginning of the unlearning process, when the student model is initialized with the weights of the $f_\text{orig}$, the student's accuracy on $D_f$ exceeds the teacher's accuracy on unseen data, since the $D_f$ comprises training data; therefore, ${\tau_d\!>\!1}$ and the teacher's output probabilities are smoothed to increase the entropy in the output space and induce uncertainty in the student model.
\smallcircle{2} As the unlearning process continues, LoTUS converges to the unlearning objective~\cref{eq:acc_objective}, the accuracy difference becomes smaller, and uncertainty is introduced with smaller steps, proportional to this accuracy difference, facilitating smooth convergence.
\smallcircle{3} If, during the unlearning process, the entropy in the student's output probabilities exceeds the desired level, then the student's accuracy on the forget set decreases below the teacher's accuracy on unseen data; therefore, $\tau_d\!<\!1$ and the teacher's output probabilities are sharpened to restore the entropy in the student's output probabilities to the desired level.
In ~\cref{fig:tau}, we illustrate how the dynamically adjusted temperature $\tau_d$ contributes to the convergence of the unlearning objective in \cref{eq:acc_objective}.
\looseness=-1

%
%
%

Statistical noise ${g\!\sim\!\text{Gumbel}(0,1)}$ added to the teacher's logits also contributes to the unlearning process. While smoothing the output probabilities does not typically alter the prediction outcome, the stochasticity introduced by $g$ facilitates the student model $f_\text{un}$ to produce predictions that differ from those of the well-converged and accurate teacher model $f_\text{orig}$. This reduces the student's accuracy on the forget set $D_f$ and drives convergence towards the objective in \cref{eq:acc_objective}. This observation aligns with the ablation analysis of Gumbel-Softmax \emph{vs.} Softmax in \cref{sec:gumbel}. 
\looseness=-1

%

To this end, the loss function in LoTUS, which guides the student model $f_\text{un}$ to align with the perturbed output probabilities of the teacher model $f_\text{orig}$, is defined for a single  instance $x$ as follows:
\begin{equation}\label{eq:loss}
\begin{split}
\ell(x,f_\text{orig},f_\text{un}) = \underbrace{l \cdot gs\big(f_\text{orig}(x), \tau_d \big) \odot \log s \big(f_\text{un}(x)\big)}_{\substack{\text{forget}}} \\
+ \underbrace{(1-l) \cdot gs\big(f_\text{orig}(x), \tau \!\to\! 0^+\big) \odot \log s\big(f_\text{un}(x)\big)}_{\substack{\text{retain}}}
\end{split}
\end{equation}
where $l\in\{0,1\}$ is an unlearning label, similar to~\cite{chundawat2023can}, indicating whether the instance belongs to the forget set $D_f$ or the retain set $D_r$, $gs(\cdot)$ is the Gumbel-Softmax function as in~\cref{eq:probabilities}, and ${s(\pi)\!=\! \exp(\pi_i) / \sum_{j=1}^k \exp(\pi_j)}$ is the Softmax function for $i=1,\dots,k$, where $k$ is the total number of classes.
For forget samples, temperature $\tau_d$ is dynamically scheduled, as shown in~\cref{eq:tau_d}, while for retain samples, $\tau$ is assigned a near-zero value to sharpen the teacher's output distribution to the greatest extent, decreasing the entropy, and enhancing retention.
\looseness=-1
\section{Experimental Setup}
We focus on the instance-wise unlearning task, while in \cref{sec:class}, we propose a LoTUS adaptation to the class unlearning task.
%
The forget sets consists of $10\%$ or $50\%$ of the training data, following~\cite{fan2024salun}.
LoTUS uses only $30\%$ of the remaining training samples as the retain set to evaluate its robustness in scenarios with limited data access.
%
To emphasize real-world conditions, we also evaluate unlearning a small portion of a large-scale dataset while restricting access to the original training data, making retraining from scratch infeasible.
To assess unlearning performance under these constraints, we introduce the novel \textit{Retrain-Free Jensen-Shannon Divergence (RF-JSD)} metric.
\looseness=-1

\noindent\textbf{Data.} Following~\cite{kurmanji2024towards, chen2024fast, tarun2023fast, chundawat2023can}, we use the CIFAR-10/100 datasets~\cite{krizhevsky2009learning}, which consist of $50,000$ training samples across 10 and 100 classes, respectively.
%
Moreover, we use the domain-specific MUFAC dataset~\cite{choi2023towards} with $8$ classes and fixed forget/retain splits. After cleaning MUFAC (see \cref{sec:mufac}), the forget set consists of ${\sim\!16\%}$ of the training data.
%
Additionally, we test TinyImageNet~\cite{le2015tiny}, which contains $100,000$ images of 200 classes and exhibits more complex data statistics than CIFAR-10/100, to further validate --beyond MUFAC-- that the assumption of distribution similarity between the forget and unseen sets in \cref{sec:upper_bound} can be relaxed. To reinforce this finding, we include an experiment with the CIFAR-10 and CIFAKE~\cite{bird2024cifake} datasets, where the unseen set is not the validation set of CIFAR-10 but consists of synthetic AI-generated data from CIFAKE.
For large-scale unlearning, we use the ImageNet1k dataset~\cite{imagenet15russakovsky} which contains $\sim\!1.2$M training samples of $1,000$ classes.
Following~\cite{qu2024learn}, we split the training set into forget/retain sets in a stratified manner to ensure robust evaluation.
\looseness=-1

\noindent\textbf{Evaluation Metrics.} Following~\cite{fan2024salun,foster2024fast, chundawat2023can} we evaluate the unlearning methods based on how closely they approximate the gold standard model, in terms of MIA accuracy and accuracy on the forget/retain/test sets, using the Average (Avg) Gap metric~\cite{fan2024salun}:
\looseness=-1
\begin{equation*}
    \text{Avg Gap} = \frac{1}{4} (|\Delta{\text{Acc}_{\text{MIA}}}| + |\Delta{\text{Acc}_f}| + |\Delta{\text{Acc}_r}| + |\Delta{\text{Acc}_t}|)
    \label{eq:avg_gap_metric}
\end{equation*}
where $|\Delta{\text{Acc}}|$ is the absolute difference in accuracy between the the unlearned and gold standard models, $\text{Acc}_\text{MIA}$ is the accuracy of the Membership Inference Attack used in~\cite{foster2024fast, chundawat2023can}, and $\text{Acc}_f$, $\text{Acc}_r$, $\text{Acc}_t$ are the accuracies on the forget, retain, and test sets, respectively. Small $\Delta{\text{Acc}_\text{MIA}}$ and $\Delta{\text{Acc}_f}$ indicate effective unlearning
while small $\Delta{\text{Acc}_r}$ and $\Delta{\text{Acc}_t}$ suggest effective retention.
Thus, Avg Gap reflects the balance between forgetting and retention.

Following~\cite{chundawat2023can}, we use the Jensen-Shannon Divergence (JSD) to further assess unlearning effectiveness and resilience to the Streisand Effect (\ie, when unlearning unintentionally makes forget samples more identifiable to attackers). The JSD provides a more sensitive measure than accuracy, as it captures distributional differences between the outputs of the unlearned and gold standard models:
\looseness=-1
%
\begin{equation*}
\begin{gathered}
    \mathcal{JS}\big(f_\text{un}(D_f) \mid\mid f_\text{gold}(D_f)\big) = \\ 
    \frac{1}{|D_f|} \sum_{x \in D_f} \Big(0.5 \cdot \mathcal{KL}\big(f_\text{un}(x) \!\mid\mid\! m\big) \!+\! 0.5 \cdot \mathcal{KL}\big(f_{gold}(x) \!\mid\mid\! m \big)\Big)
\end{gathered}
\end{equation*}
where $\mathcal{JS}$ is the Jensen-Shannon divergence~\cite{lin1991divergence}, $\mathcal{KL}$ is the Kullback-Leibler divergence~\cite{kullback1951information}, $|D_f|$ is the number of samples in the forget set,  $f_\text{un}(x)$ and $f_\text{gold}(x)$ are the predicted probability distributions for a sample $x$,
and $m$ is their average, defined as $m \!=\! \big( f_\text{un}(x) + f_\text{gold}(x) \big) / 2$.

Also, we introduce the novel Retrain-Free Jensen-Shannon Divergence (RF-JSD) metric, which does not rely on the gold standard model $f_\text{gold}$,  making it useful in real-world scenarios where model retraining is impractical or infeasible.
RF-JSD is computed by first averaging the predicted probability distributions per class from the unlearned model on the forget set and the pre-trained model on the unseen set, then averaging the JSD values between the normalized class-wise mean distributions of these models:
\looseness=-1
\begin{equation*}
\begin{gathered}
    \mathcal{JS}\big(f_\text{un}(D_f) \!\mid\mid\! f_\text{orig}(D_u)\big) = \frac{1}{k} \sum_{c=1}^{k} \mathcal{JS}(P_i \mid\mid Q_i)\\
     P_i \!=\! \frac{1}{Z_P} \sum_{j=1}^{n_i} f_\text{un}(x_j \!\mid\! y_j\!=\!i)\,,\;
    Q_i \!=\! \frac{1}{Z_Q} \sum_{j=1}^{n_i} f_\text{orig}(x_j \!\mid\! y_j\!=\!i)
\end{gathered}
\end{equation*}
where $P_i$ and $Q_i$ are the normalized class-wise mean distributions for the class $i$,
$k$ is the total number of classes, $n_i$ is the number of samples in class $i$, and $Z_P$, $Z_Q$ are sums of the mean class probabilities used for L1-normalization, ensuring that $P$, $Q$ are valid probability distributions.
\looseness=-1



RF-JSD provides greater interpretability than the retrain-free ZRF score~\cite{chundawat2023can} by aligning with the well-established JSD and maintaining a consistent optimal value of zero across different models, datasets, and forget sets. Additionally, RF-JSD is more computationally efficient, as it avoids the need for an extra randomly initialized model to establish a reference score, unlike ZRF.
\looseness=-1

\noindent\textbf{Models and Training.}
We use Vision Transformer~\cite{dosovitskiy2020image} and ResNet18~\cite{he2016deep} architectures.
Unlearning runs for $3$ epochs in ViT models and $10$ epochs in ResNet18 models, as in \cite{chundawat2023can}.
We use the AdamW optimizer with a weight decay of $5\!\times\!10^{-4}$.
Learning rates are set to $10^{\texttt{-}6}$ for ViT and $10^{\texttt{-}4}$ for ResNet18.
We perform minimal hyperparameter tuning, only on $\alpha$ in~\cref{eq:tau_d} via a search over $\{2,4,8,16\}$ to minimize the Avg Gap score without using the test set, as in~\cite{foster2024fast}; the optimal value is $\alpha\!=\!2$.
For baselines, we use the hyperparameters specified in the original papers. Baseline and hyperparameters descriptions are provided in the Supp. Material. Batch sizes remain consistent across all methods. Each experiment is evaluated using three seeds, which are also used to sample various forget sets.
\looseness=-1

\begin{table*}[h!]
\small
\centering
\begin{adjustbox}{width=\textwidth}
\begin{tabular}{l@{\hspace{1ex}}l@{\hspace{1ex}}c|ccccccccc>{\columncolor{blue!10}}c}
 && Metric ($\downarrow$)
 & Gold Std
 & Finetuning
 & NegGrad+ \cite{kurmanji2024towards}
 & RndLbl \cite{graves2021amnesiac}
 & BadT \cite{chundawat2023can}
 & SCRUB \cite{kurmanji2024towards}
 & SSD \cite{foster2024fast}
 & UNSIR \cite{tarun2023fast}
 & SalUn \cite{fan2024salun}
 & \tb{LoTUS} \\\toprule
 \multirow{9}{*}{\rotatebox[origin=c]{90}{Vision Transformer (ViT)}}
 &\multirow{3}{*}{\rotatebox[origin=c]{90}{TinyIN}}
    & Avg Gap & 0.0000 & 0.0175 & 0.0400 & 0.2925 & 0.0775 & 0.0225 & 0.0225 & 0.0225 & 0.0925 & \tb{0.0150} \\
    && JSD $\!\times\!1e4$ & 0.00\std{0.00} & 0.05\std{0.00} & 0.10\std{0.00} & 0.64\std{1.03} & 0.18\std{0.01} & 0.04\std{0.00} & 0.04\std{0.00} & 0.06\std{0.00} & 0.25\std{0.59} & \tb{0.03\std{0.00}} \\
    && Time (min.) & 228.9\std{6.49} & 22.64\std{0.02} & 25.20\std{0.02} & 25.19\std{0.02} & 16.91\std{0.05} & 33.25\std{0.01} & 27.27\std{0.06} & 21.17\std{0.22} & 76.97\std{1.72} & \tb{13.41\std{0.04}} \\
    \cmidrule{2-13}
 &\multirow{3}{*}{\rotatebox[origin=c]{90}{C-100}}
    & Avg Gap & 0.0000 & 0.0275 & 0.0325 & {0.0175} & 0.0375 & 0.0200 & {0.0175} & 0.0250
    & 0.0200
    & \tb{0.0125} \\
    && JSD $\!\times\!1e4$ & 0.00\std{0.00} & 0.07\std{0.00} & 0.13\std{0.01} & 0.06\std{0.00} & 0.17\std{0.01} & \tb{0.04\std{0.00}} & \tb{0.04\std{0.00}} & 0.08\std{0.01}
    & 0.06\std{0.01}
    & \tb{0.04\std{0.02}}\\
    && Time (min.) & 112.25\std{0.13} & 11.35\std{0.00} & 12.63\std{0.01} & 12.79\std{0.02} & {9.18\std{0.27}} & 16.74\std{0.03} & 13.67\std{0.02} & 10.69\std{0.01}
    & 38.15\std{0.04}
    & \tb{7.02\std{0.01}} \\
    \cmidrule{2-13}
 &\multirow{3}{*}{\rotatebox[origin=c]{90}{MUFAC}}
    & Avg Gap & 0.0000 & {0.0400} & 0.0475 & \tb{0.0200} & 0.1750 & \tb{0.0200} & \tb{0.0200} & 0.0475
    & \tb{0.0200}
    & \tb{0.0200} \\
    && JSD $\!\times\!1e4$ & 0.00\std{0.00} & 0.27\std{0.02} & 0.39\std{0.04} & 0.35\std{0.09} & 1.89\std{1.01} & \tb{0.05\std{0.02}} & {0.17\std{0.17}} & 0.85\std{0.06}
    & 0.29\std{0.01}
    & \tb{0.05\std{0.01}}\\
    && Time (min.)
    & 13.83\std{0.01}
    & {1.40\std{0.01}}
    & 1.67\std{0.00}
    & 1.76\std{0.01}
    & 2.09\std{0.25}
    & 2.21\std{0.01}
    & 1.91\std{0.00}
    & 3.21\std{0.01}
    & 8.15\std{0.01}
    & \tb{1.09\std{0.00}} \\
    \midrule
\multirow{9}{*}{\rotatebox[origin=c]{90}{ResNet18 (RN18)}}
 &\multirow{3}{*}{\rotatebox[origin=c]{90}{TinyIN}}
    & Avg Gap & 0.0000 & 0.2200 & 0.2250 & 0.1925 & 0.2850 & 0.2725 & 0.2700 & 0.2375 & 0.2025 & \tb{0.1675} \\
    && JSD $\!\times\!1e4$ & 0.00\std{0.00} & 1.80\std{0.04} & 1.82\std{0.07} & 1.71\std{0.11} & 1.81\std{0.04} & 0.98\std{0.00} & 0.96\std{0.02} & 1.76\std{0.05} & 1.93\std{0.09} & \tb{0.62\std{0.01}} \\
    && Time (min.) & 46.81\std{0.57} & 2.85\std{0.00} & 3.17\std{0.00} & 3.44\std{0.01} & 1.91\std{0.01} & 3.97\std{0.00} & 3.47\std{0.00} & 5.00\std{0.01} & 6.06\std{0.06} & \tb{1.62\std{0.00}} \\
    \cmidrule{2-13}
&\multirow{3}{*}{\rotatebox[origin=c]{90}{C-100}}
    & Avg Gap & 0.0000 & 0.3600 & 0.3575 & 0.4025 & 0.3675 & {0.1650} & 0.2125 & 0.3625 & 0.3650 & \tb{0.1200} \\
    && JSD $\!\times\!1e4$ & 0.00\std{0.00} & 6.88\std{0.59} & 6.87\std{0.62} & 5.84\std{0.98} & 4.30\std{0.49} & {1.87\std{0.08}} & 3.04\std{1.55} & 3.05\std{0.32} & 6.50\std{0.60} & \tb{1.67\std{0.37}} \\
    && Time (min.)
    & 3.39\std{0.30}
    & 0.43\std{0.00}
    & 0.49\std{0.00}
    & 0.57\std{0.00}
    & {0.34\std{0.01}}
    & 0.58\std{0.00}
    & 0.54\std{0.00}
    & 0.45\std{0.00}
    & 1.55\std{0.01}
    & \tb{0.30\std{0.01}} \\
    \cmidrule{2-13}
&\multirow{3}{*}{\rotatebox[origin=c]{90}{MUFAC}}
    & Avg Gap
    & 0.0000 & 0.1525 & 0.1550 & 0.1300 & \tb{0.1025} & 0.1625 & 0.1600 & 0.1450 & 0.1400 & {0.1250} \\
    && JSD $\!\times\!1e4$
    & 0.00\std{0.00}
    & 19.52\std{6.23} & 19.16\std{5.31} & 9.51\std{2.39} & 9.41\std{0.04} & {10.53\std{2.31}} & 10.30\std{2.28} & 16.32\std{4.82} & 15.25\std{5.20} & \tb{6.90\std{1.49}} \\
    && Time (min.)
    & 7.34\std{0.77}
    & 0.76\std{0.00}
    & 0.91\std{0.00}
    & 1.06\std{0.00}
    & {0.66\std{0.00}}
    & 1.20\std{0.00}
    & 1.07\std{0.00}
    & 1.68\std{0.02}
    & 2.72\std{0.02}
    & \tb{0.62\std{0.00}} \\
    \bottomrule
\end{tabular}
\end{adjustbox}
\caption{\textbf{Performance Summary of unlearning 10\% of Tiny-ImageNet (TinyIN), CIFAR-100 (C-100), and MUFAC training sets:} LoTUS outperforms state-of-the-art approaches in \tb{balancing forgetting and retention} (measured by Avg Gap), \tb{unlearning effectiveness and resilience to the Streisand effect} (indicated by JSD), and \tb{efficiency} (reflected in Time, measured in minutes).\looseness=-1}
\label{tab:avg_gap}
\end{table*}

\begin{table*}
\small
\centering
\begin{adjustbox}{width=\textwidth}
\begin{tabular}{l@{\hspace{1ex}}l@{\hspace{1ex}}|ccccccccc>{\columncolor{blue!10}}{c}>{\columncolor{green!10}}{c}}
 & Metric ($\downarrow$)
 & Gold Std
 & Finetuning
 & NegGrad+
 & RndLbl
 & BadT
 & SCRUB
 & SSD
 & UNSIR
 & SalUn
 & \tb{LoTUS} 
 & \tb{\makecell{LoTUS\\synthetic $D_u$}}\\\toprule
 \multirow{3}{*}{\rotatebox[origin=c]{90}{ViT}}
    & Avg Gap & 0.0000 & \ul{0.0075} & 0.0125 & 0.0125 & 0.0375 & \tb{0.0050} & {0.0075} & 0.0100
    & 0.0125
    & \tb{0.0050}
    & \ul{0.0075}\\
    & JSD $\!\times\!1e4$ & 0.00\std{0.00} & \tb{0.01\std{0.00}} & 0.03\std{0.00} & \ul{0.02\std{0.01}} & 0.12\std{0.03} & \tb{0.01\std{0.00}} & \ul{0.02\std{0.01}} & \tb{0.01\std{0.01}}
    & \tb{0.01\std{0.01}}
    & \tb{0.01\std{0.00}}
    & \tb{0.01\std{0.00}}\\
    & Time (min.)
    & 111.00\std{1.99}
    & 11.33\std{0.03}
    & 12.61\std{0.03}
    & 12.78\std{0.01}
    & {8.97\std{0.02}}
    & 16.66\std{0.02}
    & 13.65\std{0.02}
    & 10.68\std{0.02}
    & 37.97\std{0.19}
    & \ul{7.34\std{0.19}}
    & \tb{7.25\std{0.06}}\\
    \midrule
\multirow{3}{*}{\rotatebox[origin=c]{90}{RN18}}
    & Avg Gap
    & 0.0000 & 0.1375 & 0.0975 & 0.0925 & 0.2650 & {0.0750} & 0.0825 & 0.1075 & 0.1800 & \ul{0.0350}
    & \tb{0.0300}\\
    & JSD $\!\times\!1e4$
    & 0.00\std{0.00} & 1.03\std{0.24} & 1.06\std{0.21} & 1.00\std{0.26} & 2.39\std{2.03} & \ul{0.41\std{0.09}} & 0.82\std{0.57} & 0.65\std{0.05}
    & 1.09\std{0.05}
    & \tb{0.32\std{0.04}}
    & \tb{0.32\std{0.03}}\\
    & Time (min.)
    & 5.32\std{1.18}
    & 0.43\std{0.00}
    & 0.49\std{0.00}
    & 0.57\std{0.00}
    & {0.33\std{0.00}}
    & 0.58\std{0.00}
    & 0.54\std{0.00}
    & 0.45\std{0.00}
    & 1.56\std{0.01}
    & \tb{0.29\std{0.00}}
    & \ul{0.30\std{0.01}}\\
    \bottomrule
\end{tabular}
\end{adjustbox}
\caption{\textbf{Unlearning 10\% of CIFAR-10}. LoTUS outperforms state-of-the-art approaches, both when the calibration/unseen set $D_u$ consists of \tb{real data} ($\color{blue!70}{\bullet}$) and when it consists of \tb{synthetic data} ($\color{green!70}{\bullet}$) from the CIFAKE dataset. We highlight the \tb{best} and \ul{second-best} scores.\looseness=-1}
\label{tab:cifar}
\end{table*}
\section{Results \& Discussion}\label{sec:results}

\paragraph{Unlearning Effectiveness}
was assessed using the Avg Gap and JSD scores. Avg Gap incorporates knowledge from the MIA accuracy and the model accuracies on the forget, retain and test sets; thus it indicates the balance of forgetting/retention. 
JSD evaluates the unlearning effectiveness and the resilience to the Streisand effect.
As shown in \cref{tab:avg_gap,tab:cifar,tab:fifty}, LoTUS outperforms state-of-the-art methods in balancing forgetting/retention, unlearning effectiveness, and resilience to the Streisand effect.
As shown in \cref{tab:avg_gap}, MUFAC \& ResNet18 is the only benchmark where LoTUS succeeds the second-best and not the best Avg Gap, however MUFAC is a challenging dataset as seen by the increased JSD scores accross all methods compared to other datasets. This may derive from the increased similarity of images in the retain and forget sets, as presented in \cref{sec:failure}. 
Regarding the assumption of distributional similarity between the forget and unseen sets, in \cref{tab:cifar}, we demonstrate that it can be relaxed by showing that LoTUS is still the best-performing method even when the unseen set consists of AI-generated synthetic data from CIFAKE~\cite{bird2024cifake}.
Another intriguing finding is that across all datasets and models, LoTUS consistently achieves the highest JSD score.
\looseness = -1

The JSD metric provides a more sensitive measure of unlearning effectiveness than model's accuracy on the forget set $\text{Acc}_f$, enabling it to capture unlearning misconceptions that may lead to the Streisand effect. 
Specifically, JSD evaluates shifts in output distributions, while $\text{Acc}_f$ considers only the predicted class.
In Machine Unlearning applications, the accuracy of the pre-trained model on the forget set is typically higher than that of the gold standard model. Thus, $\text{Acc}_f$ is commonly used to assess whether unlearning reduces the pre-trained model's accuracy to align with the gold standard.
However, as emphasized by Chundawat \etal~\cite{chundawat2023can}, misclassification alone does not imply successful unlearning. They highlight a strawman unlearning solution where predictions on the forget set are maximally incorrect (\eg, a \textit{cat} is classified into the \textit{airplane} class with increased confidence), arguing that this undermines the generalization capacity of the model and increases the risk of the Streisand effect --making the forget samples more noticeable to attackers. 
The JSD score penalizes these maximally wrong predictions, while accuracy on the forget set $\text{Acc}_f$ does not. Therefore, JSD captures both unlearning effectiveness and the vulnerability to the Streisand effect, while $\text{Acc}_f$ may present misleading results.
In \cref{sec:extended_analysis}, we provide an extended analysis on how LoTUS succeeds effective unlearning on the JSD, while maintaining high accuracy even on the forget set.
Also in \cref{sec:entropy_based}, we examine the Streisand effect using an entropy-based approach as in~\cite{golatkar2020eternal}.
\looseness=-1

\vspace{-1em}
\paragraph{Unlearning Efficiency}
was assessed based on the execution time of each algorithm.
As shown in~\cref{tab:avg_gap,tab:cifar,tab:fifty,tab:imagenet}, LoTUS consistently outperforms the state-of-the-art approaches in terms of unlearning efficiency.
The time complexity of unlearning methods can be analyzed in terms of two factors: the complexity of model updates and the complexity of the auxiliary computations (such as $\tau_d$ in~\cref{eq:tau_d}).
With respect to the time complexity of model updates, the main advantage of LoTUS over Finetuning, NegGrad+, Rnd Labeling, and SCRUB is that LoTUS can preserve the model's utility using only a small percentage of retain samples, while others cannot. Considering the remaining approaches, LoTUS is more efficient mainly because it is the only one with auxiliary computations of linear complexity. A detailed analysis is presented in \cref{sec:time}.
\looseness=-1

\begin{table}
\small
\centering
\begin{adjustbox}{width=0.98\columnwidth}
\begin{tabular}{@{}l@{\hspace{1ex}}@{}l@{\hspace{1ex}}c|ccccc>{\columncolor{blue!10}}c}
 && Metric ($\downarrow$) & BadT & SCRUB & SSD & UNSIR & SalUn & \tb{LoTUS} \\\toprule
 \multirow{6}{*}{\rotatebox[origin=c]{90}{Vision Transformer}} &
 \multirow{3}{*}{\rotatebox[origin=c]{90}{C-100}}
    & Avg. Gap
    & 0.0575 & {0.0350} & {0.0350} & 0.0375 & 0.0375 & \tb{0.0225} \\
    && JSD $\!\times\! 1e4$
    & 0.06\std{0.01} & \tb{0.01\std{0.00}} & \tb{0.01\std{0.00}} & {0.02\std{0.00}} & 0.10\std{0.01} & \tb{0.01\std{0.00}} \\
    && Time (min)
    & {15.04\std{0.03}} & 16.82\std{0.03} & 18.69\std{0.06} & 18.33\std{0.02} & 38.08\std{0.02} & \tb{13.79\std{0.02}}\\
    \cmidrule{2-9}
 &\multirow{3}{*}{\rotatebox[origin=c]{90}{C-10}}
    & Avg. Gap
    & 0.0600 & {0.0125} & 0.0150 & 0.0150 & \tb{0.0050} & \tb{0.0050} \\
    && JSD $\!\times\! 1e4$
    & {0.04\std{0.01}} & \tb{0.00\std{0.00}} & \tb{0.00\std{0.00}} & \tb{0.00\std{0.00}} & 0.02\std{0.00} & \tb{0.00\std{0.00}} \\ 
    && Time (min)
    & {15.10\std{0.20}} & 16.99\std{0.35} & 19.03\std{0.54} & 18.33\std{0.02} & 37.93\std{0.18} & \tb{14.09\std{0.53}}\\
    \midrule
 \multirow{6}{*}{\rotatebox[origin=c]{90}{ResNet18}} &
 \multirow{3}{*}{\rotatebox[origin=c]{90}{C-100}}
    & Avg. Gap
    & 0.3050 & {0.2225} & {0.2225} & 0.2925 & 0.3300 &  \tb{0.1725} \\
    && JSD $\!\times\! 1e4$
    & 0.55\std{0.04} & {0.44\std{0.02}} & {0.44\std{0.02}} & 0.65\std{0.23} & 6.25\std{0.45} & \tb{0.28\std{0.00}} \\
    && Time (min)
    & 0.58\std{0.01} & 0.62\std{0.00} & 1.29\std{0.03} & 0.72\std{0.01} & 1.49\std{0.01} & \tb{0.57\std{0.01}} \\
    \cmidrule{2-9}
 & \multirow{3}{*}{\rotatebox[origin=c]{90}{C-10}}
    & Avg. Gap
    & \tb{0.0625} & 0.1075 & 0.1025 & 0.1025 & 0.1300 & {0.0650} \\
    && JSD $\!\times\! 1e4$
    & 0.18\std{0.02} & 0.14\std{0.00} & {0.13\std{0.00}} & 0.21\std{0.05} & 1.40\std{0.02} &\tb{0.09\std{0.01}}\\
    && Time (min)
    & \tb{0.57\std{0.02}} & 0.61\std{0.02} & 1.27\std{0.02} & 0.73\std{0.00} & 1.49\std{0.01} & \tb{0.57\std{0.00}}\\
    \bottomrule
\end{tabular}
\end{adjustbox}
\caption{\textbf{Scaling up the Forget set to 50\% of the training sets:} LoTUS outperforms state-of-the-art approaches in all metrics. Basic unlearning methods (Finetuning, NegGrad+, Rnd Labeling) are more efficient, but less effective than LoTUS.} 
\label{tab:fifty}
\end{table}

\vspace{-1em}
\paragraph{Large-scale unlearning on ImageNet1k.}
We consider an experimental setup that includes a ViT trained on ImageNet1k (${\sim\!1.2}$M training samples) and data access constraints that define a retain set of $45,000$ samples and forget/validation/test sets of $5,000$ samples each. The size of the ImageNet1k dataset deters retraining the model entirely to effectively unlearn the forget samples. Furthermore, when the original training dataset is not fully accessible, retraining is infeasible. This leaves Machine Unlearning as the only viable solution for removing the influence of the forget samples from the pre-trained model. Moreover, since the gold standard model is not available, the established Avg Gap and JSD metrics cannot be used. To address this, we use the RF-JSD evaluation metric, which does not require the retrained model, and has been proved to have a strong correlation with the established JSD metric. As shown in \cref{tab:pcc}, the mean Pearson correlation coefficient (PCC) of JSD and RF-JSD is $0.92\std{0.04}$ (p-value: $0.001$). As shown in~\cref{tab:imagenet}, LoTUS outperforms state-of-the-art approaches in terms of both unlearning effectiveness and efficiency.
\looseness=-1


\begin{table}
\small
\centering
\begin{adjustbox}{width=\columnwidth}
\begin{tabular}{l|cc|cc}
{Method} & {RF-JSD  $\!\times\! 1e4$ $(\downarrow)$} & {Time $(\downarrow)$} & {Retain Acc.} & {MIA Acc.} \\\toprule
Original
& 1.22\std{0.01}
& (pre-trained)             
& 0.94\std{0.00} 
& 0.71\std{0.00} \\
Finetuning
& 2.22\std{0.02}
& 16.24\std{0.03}
& 0.97\std{0.00}
& 0.78\std{0.00} \\
NegGrad+      
& 2.17\std{0.02}
& 18.10\std{0.03}
& 0.97\std{0.00}
& 0.80\std{0.00} \\
Rnd Labeling   
& 1.80\std{0.09}
& 19.37\std{0.03}
& 0.95\std{0.01}
& 0.74\std{0.01} \\
Bad Teacher   
& 3.16\std{3.25}
& 11.66\std{0.03}
& 0.77\std{0.21}
& 0.52\std{0.18} \\
SCRUB         
& 1.24\std{0.01}
& 24.49\std{0.03}
& 0.94\std{0.00}
& 0.71\std{0.00} \\
SSD           
& 1.23\std{0.01}
& 22.61\std{0.10}
& 0.94\std{0.00}
& 0.71\std{0.00} \\
UNSIR          
& 2.54\std{0.03}
& 33.12\std{0.03}
& 0.99\std{0.00}
& 0.77\std{0.01} \\
SalUn
& 1.83\std{0.03}
& 59.27\std{0.37}
& 0.95\std{0.00}
& 0.74\std{0.01} \\
\rowcolor{blue!10} \tb{LoTUS}
& \tb{1.11\std{0.01}}
& \tb{10.72\std{0.01}}
& 0.94\std{0.00}
& 0.61\std{0.01} \\
\bottomrule
\end{tabular}
\end{adjustbox}
\caption{\textbf{Large-Scale Unlearning with ImageNet1k:} LoTUS outperforms state-of-the-art approaches in both unlearning effectiveness (RF-JSD) and efficiency (Time). While other metrics lack concrete validation due to the absence of a Gold Standard, they provide additional insights: LoTUS uniquely preserves the Retain Accuracy of the pre-trained model while reducing MIA Accuracy.}
\label{tab:imagenet}
\end{table}
\section{Conclusions}
We introduced an information-theoretic framework for unlearning and proposed LoTUS, a novel method that removes the influence of specific training samples from a pre-trained model while preserving its utility on the remaining data. We demonstrated how the dynamic temperature parameter and the introduction of Gumbel noise in the activation function enable LoTUS to smooth output probabilities for forget samples, mitigating over-confident predictions that stem from data memorization.
\looseness=-1

We introduced the RF-JSD metric, which strongly correlates with the established JSD metric but eliminates the need for a retrained model, making it particularly valuable for unlearning in large-scale datasets, where retraining is impractical, or in settings with restricted data access.
We compared it with the existing ZRF score, showing that RF-JSD offers greater interpretability and efficiency.
Moreover, we highlighted that the established Avg Gap metric can produce misleading results and emphasized the increased sensitivity of JSD, which enables it to capture unlearning misconceptions that Avg Gap fails to detect.
\looseness=-1

We demonstrated that LoTUS surpasses state-of-the-art methods in both effectiveness and efficiency, demonstrating its scalability and adaptability to large-scale unlearning challenges and stringent data constraints.
\looseness=-1

\vspace{-1em}
\paragraph{Limitations.}
Both our theoretical framework and extensive experiments demonstrate that LoTUS surpasses state-of-the-art performance in instance-wise unlearning. While \cref{sec:class} shows that our theoretical framework extends to class unlearning and LoTUS can be adapted for this task, our experimental setup in class unlearning is less extensive.
\looseness=-1

\section*{Acknowledgments}
This work was partially supported by the EU funded project ATLANTIS (Grant Agreement Number 101073909).
\looseness=-1

{
    \small
    \bibliographystyle{ieeenat_fullname}
    \bibliography{main}
}

\clearpage
\setcounter{page}{1}
\maketitlesupplementary

\section{Baselines}
The \textbf{Gold Stadnard (Gold Std)} model is retrained entirely only on the retain set ($D_r$), achieving ideal unlearning --when access to the full training set is guaranteed-- but at the cost of increased computational complexity. 
\textbf{Finetuning}: The pre-trained model is further trained only on the retain samples ($D_r$).
\textbf{NegGrad+}~\cite{kurmanji2024towards}: The pre-trained model continues training on the full training set, but the gradient sign is reversed during backpropagation for the forget samples.
\textbf{Random Labeling (RndLbl)}~\cite{graves2021amnesiac}: The pre-trained model continues training on the full training set, but the forget samples are randomly reassigned to incorrect classes.
\textbf{Bad Teacher (BadT)}~\cite{chundawat2023can}: A knowledge distillation framework where the student model follows the pre-trained model for retain samples and a randomly initialized model for forget samples.
\textbf{SCRUB}~\cite{kurmanji2024towards}: A knowledge distillation framework where student model selectively aligns with the pre-trained model by minimizing the KL divergence of their outputs on retain samples while maximizing it for forget samples.
\textbf{SSD}~\cite{foster2024fast}: Weights that are disproportionately important for forget samples are identified using the Fisher Information Matrix and subsequently  dampened. 
\textbf{UNSIR}~\cite{tarun2023fast}: A noise matrix, generated based on the forget samples, is fed to the pre-trained model to maximize its error on these samples.
\textbf{SalUn}~\cite{fan2024salun}: A gradient-based approach that identifies weights to be unlearned and those to keep unchanged, followed by a downstream unlearning method such as Random Labeling.
Finetuning, NegGrad+ and Random Labeling are considered simple yet widely used unlearning baselines, whereas the latter five are state-of-the-art approaches.
\looseness=-1

LoTUS can be integrated with SalUn, with SalUn used to obtain the weight saliency mask for pruning, and LoTUS applied for unlearning. This integration can enhance the unlearning effectiveness of LoTUS. For instance, on ResNet18 with TinyImageNet, it reduces the Avg Gap of LoTUS to $0.1250$ (a $25.37\%$ decrease) and the JSD to $0.55$ (an $11.29\%$ decrease). However, this comes at the cost of efficiency, with unlearning time increasing to $4.62$ minutes (a $162\%$ increase).
\looseness=-1

\section{Reproducibility and Transparency}
The code to reproduce the results presented in this paper is publicly available at \href{https://github.com/cspartalis/LoTUS}{https://github.com/cspartalis/LoTUS}. In addition, all tables and figures have been documented in Jupyter notebooks to enhance transparency. We conducted the experiments using Python 3.11 and CUDA 12.1. For ImageNet1k experiments, we used an NVIDIA RTX A6000 48GB GPU. The remaining experiments were performed on an NVIDIA RTX 4080 16GB GPU. We also used an Intel i7-12700K CPU and 32GB RAM. The hyperparameters used for the baselines are listed in \cref{tab:hyperparams}
\looseness=-1

\begin{table}[]
    \centering
    \begin{adjustbox}{width=0.9\columnwidth}
    \begin{tabular}{l|c|c|c}
         Baseline & Learning Rate & Weight Decay & Optimizer \\\hline
         Finetune & $1 \times 10^{-3}$ & $5 \times 10^{-4}$ & SGD \\
         Negrad+ & $1 \times 10^{-3}$ & $5 \times 10^{-4}$ & SGD \\
         RndLbl & $1 \times 10^{-3}$ & $5 \times 10^{-4}$ & SGD \\
         BadT & $1 \times 10^{-4}$ & $0$ & Adam \\
         SCRUB & $5 \times 10^{-4}$ & $5 \times 10^{-4}$ & Adam \\
         SSD & $0.1$ & $0$ & SGD \\
         UNSIR & $1 \times 10^{-3}$ & $0$ & SGD \\
         SalUn & $0.1$ & $5 \times 10^{-4}$ & SGD \\
         \bottomrule
    \end{tabular}
    \end{adjustbox}
    \caption{Hyperparameters used for baselines. For state-of-the-art methods, they are taken from their respective papers.}
    \label{tab:hyperparams}
\end{table}

\section{Extended Analysis on the Accuracy Metrics}\label{sec:extended_analysis}
\begin{table*}
\small
\centering
\begin{adjustbox}{width=\textwidth}
\begin{tabular}{l@{\hspace{1ex}}l@{\hspace{1ex}}|c|c|cccccccc>{\columncolor{blue!10}}c}
 && Metric {\color{blue}($\downarrow$)} & Gold Std & Finetuning
 & NegGrad+ \cite{kurmanji2024towards}
 & RndLbl \cite{graves2021amnesiac}
 & Bad Teacher \cite{chundawat2023can}
 & SCRUB \cite{kurmanji2024towards}
 & SSD \cite{foster2024fast}
 & UNSIR \cite{tarun2023fast}
 & SalUn \cite{fan2024salun}
 & \tb{LoTUS} \\\toprule
 \multirow{20}{*}{\rotatebox[origin=c]{90}{Vision Transformer (ViT)}}
 &\multirow{5}{*}{\rotatebox[origin=c]{90}{TinyImageNet}}
    & MIA Acc.
    & 0.76\std{0.00} & 0.78\std{0.00}\bl{0.02} & 0.83\std{0.00}\bl{0.07} & 0.50\std{0.43}\bl{0.26}
    & 0.67\std{0.00}\bl{0.09} & 0.79\std{0.00}\bl{0.03}  & 0.79\std{0.00}\bl{0.03}  & 0.80\std{0.00}\bl{0.04}
    & 0.67\std{0.25}\bl{0.09}
    &  0.76\std{0.00}\tb{\bl{0.00}}\\
    && Forget Acc.
    & 0.90\std{0.00} & 0.93\std{0.00}\bl{0.03} & 0.97\std{0.00}\bl{0.07} & 0.61\std{0.52}\bl{0.29} 
    & 0.84\std{0.01}\bl{0.06} & 0.96\std{0.00}\bl{0.06} & 0.96\std{0.00}\bl{0.06} & 0.92\std{0.00}\tb{\bl{0.02}} 
    & 0.82\std{0.30}\bl{0.08}
    & 0.96\std{0.00}\bl{0.06}\\
    && Retain Acc.
    & 0.96\std{0.00} & 0.98\std{0.00}\bl{0.02} & 0.98\std{0.00}\bl{0.02} & 0.64\std{0.55}\bl{0.32}
    & 0.87\std{0.01}\bl{0.09} & 0.96\std{0.00}\tb{\bl{0.00}} & 0.96\std{0.00}\tb{\bl{0.00}} & 0.94\std{0.00}\bl{0.02} & 0.86\std{0.32}\bl{0.10}
    & 0.96\std{0.00}\tb{\bl{0.00}}\\
    && Test Acc.
    & 0.90\std{0.00} & 0.90\std{0.00}\tb{\bl{0.00}} &  0.90\std{0.00}\tb{\bl{0.00}} & 0.60\std{0.51}\bl{0.30}
    & 0.83\std{0.01}\bl{0.07} & 0.90\std{0.00}\tb{\bl{0.00}} & 0.90\std{0.00}\tb{\bl{0.00}} & 0.89\std{0.00}\bl{0.01}
    & 0.80\std{0.30}\bl{0.10}
    & 0.90\std{0.00}\tb{\bl{0.00}}\\
    && \B{Avg Gap}
    & \B{0.0000} & \B{0.0175} & \B{0.0400} & \B{0.2925} & \B{0.0775} & \B{0.0225} & \B{0.0225} & \B{0.0225} 
    & \B{0.0925}
    & \tb{\B{0.0150}} \\
    \cmidrule{2-13}
 &\multirow{5}{*}{\rotatebox[origin=c]{90}{CIFAR-100}}
    & MIA Acc.
    & 0.72\std{0.00} & 0.77\std{0.00}\bl{0.05} & 0.79\std{0.02}\bl{0.07} & 0.74\std{0.01}{\bl{0.02}}
    & 0.66\std{0.01}\bl{0.06} & 0.75\std{0.00}\bl{0.03} & 0.75\std{0.01}\bl{0.03} & 0.78\std{0.01}\bl{0.06}
    & 0.75\std{0.01}\bl{0.03}
    & 0.71\std{0.02}\tb{\bl{0.01}} \\
    && Forget Acc.
    & 0.92\std{0.00} & 0.95\std{0.01}{\bl{0.03}} & 0.97\std{0.02}\bl{0.05} & {0.94\std{0.00}\tb{\bl{0.02}}}
    & {0.90\std{0.01}\tb{\bl{0.02}}} & 0.97\std{0.00}\bl{0.05} & 0.96\std{0.01}\bl{0.04} & {0.94\std{0.00}\tb{\bl{0.02}}}
    & 0.94\std{0.01}\bl{0.02}
    & 0.96\std{0.01}\bl{0.04} \\
    && Retain Acc.
    & 0.96\std{0.00} & 0.98\std{0.00}\bl{0.02} & 0.97\std{0.02}{\bl{0.01}} & 0.98\std{0.00}\bl{0.02}
    & 0.91\std{0.00}\bl{0.05} & 0.96\std{0.00}\tb{\bl{0.00}} & 0.96\std{0.00}\tb{\bl{0.00}} & 0.95\std{0.01}{\bl{0.01}}
    & 0.98\std{0.01}\bl{0.02}
    & 0.96\std{0.00}\tb{\bl{0.00}} \\
    && Test Acc.
    & 0.91\std{0.01} & 0.92\std{0.01}{\bl{0.01}} & 0.91\std{0.01}\tb{\bl{0.00}} & 0.92\std{0.00}{\bl{0.01}}
    & 0.89\std{0.01}\bl{0.02} & 0.91\std{0.01}\tb{\bl{0.00}} & 0.91\std{0.00}\tb{\bl{0.00}} & 0.90\std{0.01}{\bl{0.01}}
    & 0.92\std{0.00}\bl{0.01}
    & 0.91\std{0.00}\tb{\bl{0.00}} \\
    && \B{Avg Gap}
    & \B{0.0000} & \B{0.0275} & \B{0.0325} & {\B{0.0175}} & \B{0.0375} & \B{0.0200} & {\B{0.0175}} & \B{0.0250} 
    & \B{0.0200}
    & \tb{\B{0.0125}} \\
    \cmidrule{2-13}
 &\multirow{5}{*}{\rotatebox[origin=c]{90}{CIFAR-10}}
    & MIA Acc.
    & 0.88\std{0.00} & 0.90\std{0.00}\bl{0.02} & 0.91\std{0.00}\bl{0.03} & 0.84\std{0.02}\bl{0.04}
    & 0.81\std{0.02}\bl{0.07} & 0.88\std{0.00}\tb{\bl{0.00}} & 0.89\std{0.01}{\bl{0.01}} & 0.90\std{0.00}\bl{0.02}
    & 0.84\std{0.02}\bl{0.04}
    & 0.87\std{0.00}{\bl{0.01}} \\
    && Forget Acc.
    & 0.99\std{0.00} & 0.99\std{0.00}\tb{\bl{0.00}} & 1.00\std{0.00}{\bl{0.01}} & 0.99\std{0.00}\tb{\bl{0.00}}
    & 0.96\std{0.01}\bl{0.03} & 1.00\std{0.00}{\bl{0.01}} & 1.00\std{0.01}{\bl{0.01}} & 0.99\std{0.00}\tb{\bl{0.00}}
    & 0.99\std{0.00}\tb{\bl{0.00}}
    & 1.00\std{0.00}{\bl{0.01}} \\
    && Retain Acc.
    & 1.00\std{0.00} & 1.00\std{0.00}\tb{\bl{0.00}} & 1.00\std{0.00}\tb{\bl{0.00}} & 1.00\std{0.00}\tb{\bl{0.00}}
    & 0.97\std{0.01}\bl{0.03} & 1.00\std{0.00}\tb{\bl{0.00}} & 1.00\std{0.01}\tb{\bl{0.00}} & 0.99\std{0.00}{\bl{0.01}}
    & 1.00\std{0.00}\tb{\bl{0.00}}
    & 1.00\std{0.00}\tb{\bl{0.00}} \\
    && Test Acc.
    & 0.98\std{0.01} & 0.99\std{0.01}\bl{0.01} & 0.99\std{0.01}\bl{0.01} & 0.99\std{0.00}\bl{0.01}
    & 0.96\std{0.01}\bl{0.02} & 0.99\std{0.01}\bl{0.01} & 0.99\std{0.01}\bl{0.01} & 0.99\std{0.00}\bl{0.01}
    & 0.99\std{0.01}\bl{0.01}
    & 0.98\std{0.01}\tb{\bl{0.00}} \\
    && \B{Avg Gap}
    & \B{0.0000} & {\B{0.0075}} & \B{0.0125} & \B{0.0125} & \B{0.0375} & \tb{\B{0.0050}} & {\B{0.0075}} & \B{0.0100}
    & \B{0.0125}
    & \tb{\B{0.0050}} \\
    \cmidrule{2-13}
&\multirow{5}{*}{\rotatebox[origin=c]{90}{MUFAC}}
    & MIA Acc.
    & 0.57\std{0.00} & 0.52\std{0.08}{\bl{0.05}} & 0.52\std{0.07}{\bl{0.05}} & 0.52\std{0.10}{\bl{0.05}}
    & 0.35\std{0.05}\bl{0.22} & 0.59\std{0.01}\tb{\bl{0.02}} & 0.59\std{0.01}\tb{\bl{0.02}} & 0.47\std{0.08}\bl{0.10}
    & 0.53\std{0.12}\bl{0.04}
    & 0.59\std{0.01}\tb{\bl{0.02}} \\
    && Forget Acc.
    & 0.57\std{0.01} & 0.61\std{0.01}\bl{0.04} & 0.66\std{0.02}\bl{0.09} & 0.58\std{0.01}\tb{\bl{0.01}}
    & 0.43\std{0.06}\bl{0.14} & 0.62\std{0.01}\bl{0.05} & 0.59\std{0.04}{\bl{0.02}} & 0.58\std{0.01}\tb{\bl{0.01}}
    & 0.58\std{0.01}\tb{\bl{0.01}}
    & 0.63\std{0.00}\bl{0.06} \\
    && Retain Acc.
    & 0.66\std{0.01} & 0.72\std{0.01}\bl{0.06} & 0.71\std{0.01}\bl{0.05} & 0.67\std{0.02}{\bl{0.01}}
    & 0.47\std{0.07}\bl{0.19} & 0.66\std{0.01}\tb{\bl{0.00}} & 0.63\std{0.04}\bl{0.03} & 0.72\std{0.01}\bl{0.06}
    & 0.68\std{0.01}\bl{0.02}
    & 0.66\std{0.01}\tb{\bl{0.00}} \\
    && Test Acc.
    & 0.65\std{0.01} & 0.66\std{0.01}{\bl{0.01}} & 0.65\std{0.03}\tb{\bl{0.00}} & 0.64\std{0.01}{\bl{0.01}}
    & 0.50\std{0.08}\bl{0.15} & 0.66\std{0.01}{\bl{0.01}} & 0.64\std{0.01}{\bl{0.01}} & 0.63\std{0.02}\bl{0.02}
    & 0.64\std{0.02}\bl{0.01}
    & 0.65\std{0.01}\tb{\bl{0.00}} \\
    && \B{Avg Gap}
    & \B{0.0000} & {\B{0.0400}} & \B{0.0475} & \tb{\B{0.0200}} & \B{0.1750} & \tb{\B{0.0200}} & \tb{\B{0.0200}} & \B{0.0475}
    & \tb{\B{0.0200}}
    & \tb{\B{0.0200}} \\
    \midrule
\multirow{20}{*}{\rotatebox[origin=c]{90}{ResNet18 (RN18)}}
& \multirow{5}{*}{\rotatebox[origin=c]{90}{TinyImageNet}}
    & MIA Acc.
    & 0.30\std{0.01} & 0.00\std{0.00}\bl{0.30} & 0.00\std{0.00}\bl{0.30} & 0.00\std{0.00}\bl{0.30}
    & 0.67\std{0.52}\bl{0.37} & 0.96\std{0.01}\bl{0.66}  & 0.95\std{0.01}\bl{0.65}  & 0.67\std{0.58}\bl{0.37}
    &  0.00\std{0.00}\bl{0.30}
    &  0.53\std{0.01}\tb{\bl{0.23}}\\
    && Forget Acc.
    & 0.58\std{0.00} & 0.70\std{0.02}\bl{0.12}  & 0.73\std{0.02}\bl{0.15}  & 0.56\std{0.02}\tb{\bl{0.02}}
    & 0.49\std{0.04}\bl{0.09} & 1.00\std{0.00}\bl{0.42} & 1.00\std{0.00}\bl{0.42}  & 0.68\std{0.03}\bl{0.10}
    & 0.62\std{0.01}\bl{0.04}
    & 0.91\std{0.01}\bl{0.33} \\
    && Retain Acc.
    & 1.00\std{0.00} & 0.73\std{0.02}\bl{0.27} & 0.73\std{0.02}\bl{0.27}  & 0.73\std{0.02}\bl{0.27}
    & 0.55\std{0.04}\bl{0.45} & 1.00\std{0.00}\tb{\bl{0.00}}  & 1.00\std{0.00}\tb{\bl{0.00}} & 0.71\std{0.02}\bl{0.29}
    & 0.71\std{0.02}\bl{0.29}
    & 0.93\std{0.01}\bl{0.07}\\
    && Test Acc.
    & 0.89\std{0.01} & 0.40\std{0.01}\bl{0.19} & 0.41\std{0.01}\bl{0.18} & 0.41\std{0.02}\bl{0.18}
    & 0.36\std{0.03}\bl{0.23} & 0.60\std{0.01}\tb{\bl{0.01}} & 0.60\std{0.00}\tb{\bl{0.01}} & 0.40\std{0.02}\bl{0.19}
    & 0.41\std{0.01}\bl{0.18}
    & 0.55\std{0.00}\bl{0.04} \\
    && \B{Avg Gap}
    & \B{0.0000} & \B{0.2200} & \B{0.2250} & \B{0.1925} & \B{0.2850} & \B{0.2725} &  \B{0.2700} & \B{0.2375} 
    & \B{0.2025} 
    & \tb{\B{0.1675}} \\
    \cmidrule{2-13}
&\multirow{5}{*}{\rotatebox[origin=c]{90}{CIFAR-100}}
    & MIA Acc.
    & 0.49\std{0.01} & 0.00\std{0.00}\bl{0.49} & 0.00\std{0.00}\bl{0.49} & 0.00\std{0.00}\bl{0.49}
    & 0.33\std{0.58}\bl{0.16} & 0.78\std{0.05}\bl{0.29} & 0.59\std{0.05}{\bl{0.10}} & 0.00\std{0.00}\bl{0.49}
    & 0.00\std{0.00}\bl{0.49}
    & 0.28\std{0.22}\tb{\bl{0.21}} \\
    && Forget Acc.
    & 0.57\std{0.02} & 0.40\std{0.06}\bl{0.17} & 0.41\std{0.06}{\bl{0.16}} & 0.31\std{0.06}\bl{0.26}
    & 0.27\std{0.03}\bl{0.30} & 0.93\std{0.03}\bl{0.36} & 0.50\std{0.32}\tb{\bl{0.07}} & 0.40\std{0.07}\bl{0.17}
    & 0.38\std{0.04}\bl{0.19}
    & 0.81\std{0.08}\bl{0.24} \\
    && Retain Acc.
    & 0.94\std{0.03} & 0.41\std{0.06}\bl{0.53} & 0.41\std{0.06}\bl{0.53} & 0.37\std{0.07}\bl{0.57}
    & 0.28\std{0.03}\bl{0.66} & 0.93\std{0.03}\tb{\bl{0.01}} & 0.50\std{0.32}{\bl{0.44}} & 0.41\std{0.07}\bl{0.53}
    & 0.41\std{0.04}\bl{0.53}
    & 0.92\std{0.02}\tb{\bl{0.02}} \\
    && Test Acc.
    & 0.60\std{0.02} & 0.35\std{0.05}\bl{0.25} & 0.35\std{0.05}\bl{0.25} & 0.31\std{0.06}\bl{0.29}
    & 0.25\std{0.03}\bl{0.35} & 0.60\std{0.02}\tb{\bl{0.00}} & 0.36\std{0.20}\bl{0.24} & 0.34\std{0.04}\bl{0.26}
    & 0.35\std{0.03}\bl{0.25}
    & 0.61\std{0.01}{\bl{0.01}} \\
    && \B{Avg Gap}
    & \B{0.0000} & \B{0.3600} & \B{0.3575} & \B{0.4025} & \B{0.3675} & {\B{0.1650}} & \B{0.2125} & \B{0.3625}
    & \B{0.3650}
    & \tb{\B{0.1200}} \\
    \cmidrule{2-13}
&\multirow{6}{*}{\rotatebox[origin=c]{90}{CIFAR-10}}
    & MIA Acc.
    & 0.76\std{0.03} & 0.30\std{0.26}\bl{0.46} & 0.48\std{0.50}\bl{0.28} & 0.48\std{0.50}\bl{0.28}
    & 0.43\std{0.37}\bl{0.33} & 0.94\std{0.01}\bl{0.18} & 0.81\std{0.11}\tb{\bl{0.05}} & 0.46\std{0.03}\bl{0.30}
    & 0.16\std{0.28}\bl{0.60}
    & 0.82\std{0.10}{\bl{0.06}} \\
    && Forget Acc.
    & 0.91\std{0.02} & 0.97\std{0.01}\bl{0.06} & 0.97\std{0.01}\bl{0.06} & 0.96\std{0.01}{\bl{0.05}}
    & 0.71\std{0.18}\bl{0.20} & 1.00\std{0.00}\bl{0.09} & 0.86\std{0.16}{\bl{0.05}} & 0.93\std{0.01}\tb{\bl{0.02}}
    & 0.94\std{0.02}\tb{\bl{0.02}}
    & 0.99\std{0.00}\bl{0.08} \\
    && Retain Acc.
    & 0.99\std{0.02} & 0.98\std{0.01}{\bl{0.01}} & 0.97\std{0.01}\bl{0.02} & 0.97\std{0.01}\bl{0.02}
    & 0.71\std{0.18}\bl{0.28} & 1.00\std{0.00}{\bl{0.01}} & 0.87\std{0.16}\bl{0.12} & 0.93\std{0.01}\bl{0.06}
    & 0.95\std{0.02}\bl{0.04}
    & 0.99\std{0.00}\tb{\bl{0.00}} \\
    && Test Acc.
    & 0.91\std{0.02} & 0.89\std{0.02}\bl{0.02} & 0.88\std{0.02}\bl{0.03} & 0.89\std{0.02}\bl{0.02}
    & 0.66\std{0.16}\bl{0.25} & 0.93\std{0.01}{\bl{0.02}} & 0.80\std{0.15}\bl{0.11} & 0.86\std{0.01}\bl{0.05}
    & 0.86\std{0.03}\bl{0.05}
    & 0.91\std{0.01}\tb{\bl{0.00}} \\
    && \B{Avg Gap}
    & \B{0.0000} & \B{0.1375} & \B{0.0975} & \B{0.0925} & \B{0.2650} & {\B{0.0750}} & \B{0.0825} & \B{0.1075}
    & \B{0.1800}
    & \tb{\B{0.0350}} \\
    \cmidrule{2-13}
& \multirow{5}{*}{\rotatebox[origin=c]{90}{MUFAC}}
    & MIA Acc.
    & 0.48\std{0.04} & 0.54\std{0.09}{\bl{0.06}} & 0.53\std{0.08}\tb{\bl{0.05}} & 0.33\std{0.31}\bl{0.15}
    & 0.34\std{0.01}\bl{0.14} & 0.70\std{0.05}\bl{0.22} & 0.70\std{0.06}\bl{0.22} & 0.40\std{0.35}\bl{0.08}
    & 0.53\std{0.08}\tb{\bl{0.05}}
    & 0.53\std{0.04}\tb{\bl{0.05}} \\
    && Forget Acc.
    & 0.47\std{0.04} & 0.64\std{0.04}{\bl{0.17}} & 0.68\std{0.04}\bl{0.21} & 0.66\std{0.04}\bl{0.19}
    & 0.53\std{0.07}\tb{\bl{0.06}} & 0.88\std{0.06}\bl{0.41} & 0.87\std{0.06}\bl{0.40} & 0.71\std{0.03}\bl{0.24}
    & 0.63\std{0.05}\bl{0.16}
    & 0.86\std{0.04}\bl{0.39} \\
    && Retain Acc.
    & 0.89\std{0.04} & 0.64\std{0.04}\bl{0.25} & 0.66\std{0.03}\bl{0.23} & 0.80\std{0.03}\bl{0.09}
    & 0.76\std{0.04}\bl{0.13} & 0.89\std{0.04}\tb{\bl{0.00}} & 0.89\std{0.05}\tb{\bl{0.00}} & 0.73\std{0.03}\bl{0.16}
    & 0.67\std{0.04}\bl{0.22}
    & 0.85\std{0.08}{\bl{0.04}} \\
    && Test Acc.
    & 0.56\std{0.02} & 0.43\std{0.01}\bl{0.13} & 0.43\std{0.01}\bl{0.13} & 0.47\std{0.02}\bl{0.09}
    & 0.48\std{0.03}{\bl{0.08}} & 0.54\std{0.03}\tb{\bl{0.02}} & 0.54\std{0.03}\tb{\bl{0.02}} & 0.46\std{0.01}\bl{0.10}
    & 0.43\std{0.02}\bl{0.13}
    & 0.54\std{0.05}\tb{\bl{0.02}} \\
    && \B{Avg Gap}
    & \B{0.0000} & \B{0.1525} & \B{0.1550} & \B{0.1300} & \tb{\B{0.1025}} & \B{0.1625} & \B{0.1600} & \B{0.1450}
    & \B{0.1400}
    & {\B{0.1250}} \\
\bottomrule
\end{tabular}
\end{adjustbox}
\caption{\textbf{Accuracy Metrics used to compute Average (Avg) Gap}. Mean performance and standard deviation ($\mu \!\pm\! \sigma$) are reported across three trials with different forget and retain sets. Performance gaps relative to the Gold Standard are noted as \bl{$\bullet$}, with smaller gaps indicating stronger performance. Avg Gap serves as a key indicator, summarizing performance across MIA, Forget, Retain, and Test Accuracy. LoTUS achieves state-of-the-art results in MIA, retain and test accuracies, ranking as the best in most cases and second-best in the remaining.\looseness=-1}
\label{tab:avg_gap_supp}
\end{table*}
\Cref{tab:avg_gap_supp} presents the accuracy scores that define the Avg Gap metric. 
Beyond outperforming state-of-the-art methods in terms of Avg Gap, LoTUS achieves the best scores in individual accuracy metrics, including MIA accuracy, and accuracy on the retain and test sets. Specifically, it consistently ranks either first or second in these metrics, with first place being the most frequent.
\looseness=-1
 
Regarding retention performance (\ie, preserving the utility of the pre-trained model), LoTUS clearly outperforms state-of-the-art, as evidenced by its superior accuracy on the retain and test sets.
\looseness=-1
 
However, evaluating unlearning effectiveness, requires a more nuanced analysis. Although LoTUS consistently ranks among the top two methods in MIA accuracy, its accuracy on the forget set exceeds that of the gold standard model (\ie, the model retrained solely on the retain set). This apparent discrepancy may lead to misleading evaluation, suggesting that LoTUS exhibits poor unlearning performance.
\looseness=-1

However, by incorporating the more sensitive JSD metric --a measure that captures distributional-level differences and provides a more robust evaluation, as detailed in \cref{sec:results}-- we conclude that LoTUS achieves effective unlearning. Given this,  the increased accuracy on the forget set does not indicate poor unlearning, but rather suggests that LoTUS preserves the utility of the pre-trained model even for the forget samples.
\looseness=-1
The fact that LoTUS achieves the best Avg Gap scores despite the disproportionate penalty imposed by the gap between the accuracy of the unlearned and gold standard models on forget samples further reinforces its capacity to balance forgetting and retention, as evidenced by Avg Gap.
\looseness=-1

This also raises concerns about the widely used Avg Gap metric, as it may lead to misleading evaluation of unlearning. However, incorporating both Avg Gap and JSD metrics in the evaluation helps mitigate these concerns.
\looseness=-1

\section{Detailed Comparison of RF-JSD and ZRF}
The ZRF metric~\cite{chundawat2023can} assesses the unlearning effectiveness by computing the JSD score twice: once between the unlearned and a randomly initialized model, and again between the pre-trained and the same randomly initialized model. The latter serves as a reference point for the optimal value.
\looseness=-1

By contrast, RF-JSD simplifies the evaluation by requiring only a single JSD computation --between the unlearned model and the original model-- where the optimal value is fixed at zero. This direct alignment with the JSD metric (which also has an optimal value fixed at zero) facilitates a more comprehensive evaluation of the unlearning effectiveness.
\looseness=-1

Beyond the obvious efficiency gain from RF-JSD not requiring inference on an additional randomly initialized model to obtain a reference score --unlike ZRF-- its use of normalized class-wise mean distributions further enhances computational efficiency.
Specifically, this reduces the complexity from $O(n_f \cdot n_u \cdot k)$ to $O\big((n_f + n_u)\cdot k\big)$, where $n_f$ and $n_u$ denote the number of samples in the forget and test sets, respectively, and $k$ is the number of classes. This optimization significantly reduces the computational overhead, particularly for large datasets.
In this analysis, we exclude the complexity of the feed-forward process, which remains unchanged.
\looseness=-1

Finally, \Cref{tab:pcc} presents a detailed correlation between RF-JSD and JSD as measured by the Pearson correlation coefficient (PCC) for all benchmarks. PCC results exhibit a strong correlation between these two metrics, with RF-JSD offering the additional advantage of not requiring a retrained model (\ie, gold standard).
\looseness=-1
\begin{table}
\small
\centering
\begin{adjustbox}{width=1\columnwidth}
\begin{tabular}{@{}l@{\hspace{2ex}}l|cc}
 & Dataset $\Big(\frac{\text{num. of forget samples}}{\text{num. of training samples}} \times 100\% \Big)$ & PCC ($\uparrow$) & p-value ($\downarrow$) \\\toprule
\multirow{5}{*}{\rotatebox[origin=c]{90}{ViT}}
& CIFAR-100 \,\, ($10\%$) & 0.84 & 0.0043 \\
& CIFAR-10 \quad ($10\%$) & 0.92 & 0.0005 \\
& MUFAC & 0.93 & 0.0003 \\
& CIFAR-100 \,\, ($50\%$) & 0.94 & 0.0001 \\
& CIFAR-10 \quad ($50\%$) & 0.99 & 0.0000 \\
\midrule
\multirow{5}{*}{\rotatebox[origin=c]{90}{ResNet18}}
& CIFAR-100 \,\, ($10\%$) & 0.97 & 0.0000 \\
& CIFAR-10 \quad ($10\%$) & 0.90 & 0.0011 \\
& MUFAC & 0.88 & 0.0018 \\
& CIFAR-100 \,\, ($50\%$) & 0.91 & 0.0006 \\
& CIFAR-10 \quad ($50\%$) & 0.89 & 0.0013 \\
\midrule
& {Mean $\pm$ Std} & 0.92\std{0.04} & 0.0010\std{0.0016} \\
\bottomrule
\end{tabular}
\end{adjustbox}
\caption{\textbf{Retrain Free-JSD (RF-JSD) and JSD Correlation} measured with the Pearson correlation coefficient (PCC). A high PCC (closer to 1) indicates a strong correlation, while a low p-value reflects high confidence in the measurement. The table shows that RF-JSD strongly correlates with the well-established JSD metric across datasets and architectures, demonstrating its reliability as unlearning metric that is particularly useful when the gold standard model is not available (\eg, it is impractical due to high computational complexity or it is infeasible due to not access to the original training set) .\looseness=-1}
\label{tab:pcc}
\end{table}

\section{Detailed Analysis on the Time Complexity}\label{sec:time}
This section provides an in-depth analysis that demonstrates why LoTUS achieves superior efficiency compared to state-of-the-art approaches, as observed in~\cref{tab:avg_gap,tab:fifty,tab:imagenet} and discussed in ~\cref{sec:results}.
\looseness=-1
We define the time complexity of model updates in DNNs, generalized across architectures like ResNet18 and ViT, as follows:
\begin{equation}
    O(E \!\cdot\! \frac{n_f + n_r}{B} \!\cdot\! N_p \!\cdot\! N_i)
    \label{eq:complexity_supp}
\end{equation}
where $E$ represents the total number of epochs, $n_f$ and $n_r$ are the number of instances in $D_f$ (forget set) and $D_r$ (retain set) used during unlearning, respectively, $B$ is the batch size, $N_p$ is the total number of model parameters, and $N_i$ is the input dimensionality. While this definition abstracts away architectural-specific details and optimizations, it provides a meaningful framework for comparing methods on shared benchmarks.
\looseness=-1

The main advantage of LoTUS over Finetuning, NegGrad+, Random Labeling, and SCRUB is that it requires significantly fewer instances $n_r$ from the retain set $D_r$. Specifically, LoTUS can use only $30\%$ of the instances in $D_r$ to preserve the utility of the model. All other factors ($E,n_f,B,N_p,N_i$) are the same for all unlearning baselines in our benchmarks. As shown in \cref{tab:avg_gap}, LoTUS achieves superior efficiency.
\looseness=-1

As the number of instances $n_f$ in the forget set increases, the execution time of LoTUS increases, in alignment with \cref{eq:complexity_supp}. Thus, in the extreme scenario where $50\%$ of the forget set is designated for unlearning, we observe that the efficiency of Finetuning, NegGrad+, and Random Labeling may exceed that of LoTUS, as shown in \cref{tab:fifty}. In \cref{tab:fifty_supp} we present the scores of these basic unlearning methods that are not presented in \cref{tab:fifty}, and show that they may be better in terms of efficiency, but LoTUS remains the best in terms of effectiveness.
\looseness=-1
\begin{table}
\small
\centering
\begin{adjustbox}{width=\columnwidth}
\begin{tabular}{@{}l@{\hspace{1ex}}c|ccc>{\columncolor{blue!10}}c}
 & Metric ($\downarrow$) & Finetuning & NegGrad+ & RndLbl & \tb{LoTUS} \\\toprule
 \multirow{3}{*}{\rotatebox[origin=c]{90}{\makecell{ViT \\ C-100}}}
    & Avg. Gap
    & 0.0400 & 0.0600 & 0.0250 & \tb{0.0225} \\
    & JSD $\!\times\! 1e4$
    & 0.02\std{0.00} & {0.03\std{0.01}} & \tb{0.01\std{0.01}} & \tb{0.01\std{0.00}} \\
    & Time (min)
    & \tb{6.34\std{0.01}} & 12.68\std{0.02} & 12.63\std{0.02} & {13.79\std{0.02}}\\
    \midrule
 \multirow{3}{*}{\rotatebox[origin=c]{90}{\makecell{ViT \\ C-10}}}
    & Avg. Gap
    & 0.0125 & {0.0200} & \tb{0.0050} & \tb{0.0050} \\
    & JSD $\!\times\! 1e4$
    & \tb{0.00\std{0.00}} & {0.01\std{0.00}} & \tb{0.00\std{0.00}} & \tb{0.00\std{0.00}} \\ 
    & Time (min)
    & \tb{6.48\std{0.27}} & 12.97\std{0.50} & 12.60\std{0.03} & 14.09\std{0.53} \\
    \midrule
 \multirow{3}{*}{\rotatebox[origin=c]{90}{\makecell{RN18\\C-100}}}
    & Avg. Gap
    & 0.3200 & {0.3150} & {0.3875} & \tb{0.1725} \\
    & JSD $\!\times\! 1e4$
    & 1.39\std{0.10} & {1.38\std{0.08}} & {1.03\std{0.23}} & \tb{0.28\std{0.00}} \\
    & Time (min)
    & \tb{0.26\std{0.01}} & 0.52\std{0.00} & 0.48\std{0.00} & {0.57\std{0.01}} \\
    \midrule
 \multirow{3}{*}{\rotatebox[origin=c]{90}{\makecell{RN18\\C-10}}}
    & Avg. Gap
    & {0.1100} & 0.1475 & 0.2100 & \tb{0.0650} \\
    & JSD $\!\times\! 1e4$
    & 0.31\std{0.00} & 0.31\std{0.01} & {0.73\std{0.22}} & \tb{0.09\std{0.01}}\\
    & Time (min)
    & \tb{0.26\std{0.01}} & 0.51\std{0.02} & 0.48\std{0.00} & {0.57\std{0.00}}\\
    \bottomrule
\end{tabular}
\end{adjustbox}
\caption{\textbf{Scaling up the Forget set to 50\% of the training sets:} LoTUS outperforms basic unlearning methods in unlearning effectiveness, but not in efficiency.} 
\label{tab:fifty_supp}
\end{table}

Next, we compare the time complexity of the auxiliary computations between LoTUS and other unlearning baselines that use equal or fewer samples from the retain set $D_r$:
\looseness=-1
\begin{description}
\setlength\itemsep{0em}
\item{LoTUS:} ${O(n_f\!+\!n_v)}$, where $n_v$ is the total number of instances in the validation set, for computing $\tau_d$.
\looseness=-1
\item{Bad Teacher~\cite{chundawat2023can}:} ${O\big( (n_f\!+\!n_r) \!\cdot\! k \big)}$, where $k$ is the total number of classes, for calculating the $\mathcal{KL}$ divergences between the student and the teacher.
\looseness=-1
\item{UNSIR~\cite{tarun2023fast}:} ${O(E_{noise} \!\cdot\! n_f \!\cdot\! N_i)}$, where $E_{noise}$ are the epochs for noise optimization, and $N_i$ represents the total input dimensionality (product of channels, width and height of the images).
\looseness=-1
\item{SSD~\cite{foster2024fast}:} ${O(n_f \!\cdot\! N_p^2)}$ for computing the Fisher Information Matrix.
\looseness=-1
\end{description}

In this analysis, we exempt the complexity of the feed-forward process which is the same for all the unlearning methods in our benchmarks.
Also, SalUn~\cite{fan2024salun} introduces a computational overhead prior to unlearning due to the computations of the saliency mask for weight pruning. The complexity of this auxiliary computation contributes to the overall complexity of the downstream method used for unlearning (\eg, Random Labeling and LoTUS in our case).
Among the unlearning methods, LoTUS is the only one with auxiliary computations of linear complexity.
\looseness=-1

\section{Cleaning the MUFAC Dataset}\label{sec:mufac}
\begin{figure}
    \centering
    \includegraphics[width=0.7\columnwidth]{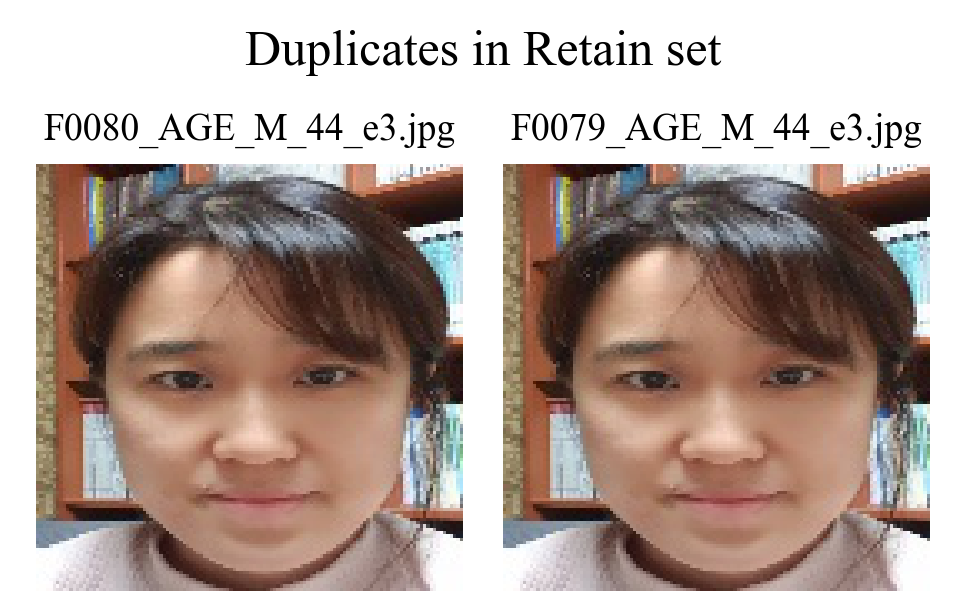}
    \includegraphics[width=0.7\columnwidth]{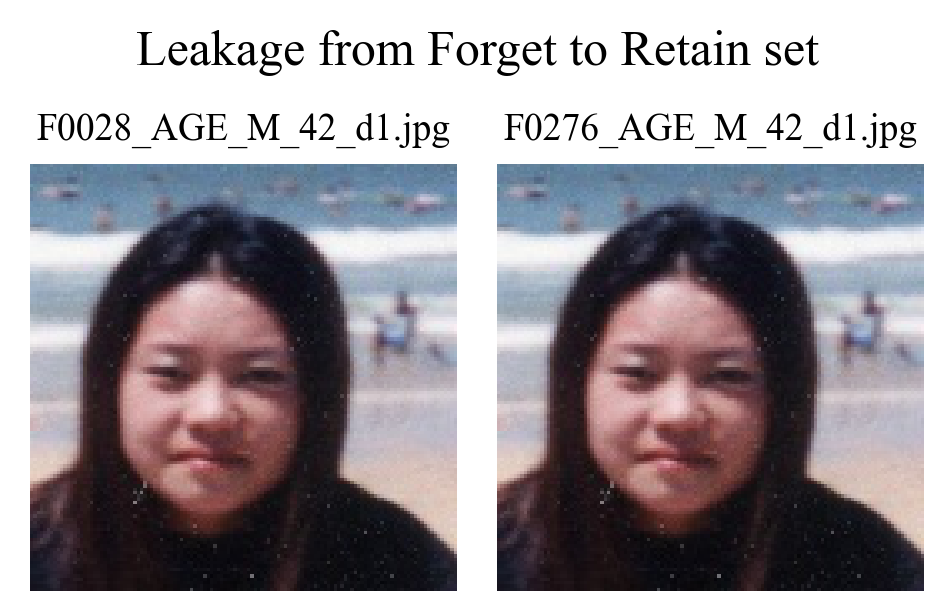}
    \vspace{-0.5em} 
\caption{\textbf{Duplicates in MUFAC:} An example of a duplicate within the retain set (top) and a critical duplicate shared between the retain and forget set (bottom), which introduces information leakage.\looseness=-1}
\label{fig:mufac_examples}
\end{figure}

We identified duplicates within the forget, retain, validation, and test splits of the MUFAC dataset. More critically, we discovered instances of information leakage across these splits.
To address this, we used image hashing to detect identical images with different filenames in these splits, as shown in \cref{fig:mufac_examples}.
\looseness=-1

After cleaning MUFAC, the retain set contains $5,513$ samples, and the forget set contains $1,062$ samples. We provide the code for identifying duplicate images and cleaning MUFAC in \href{https://github.com/cspartalis/LoTUS}{https://github.com/cspartalis/LoTUS}.
\looseness=-1

\begin{figure}
    \centering
    \includegraphics[width=0.8\columnwidth]{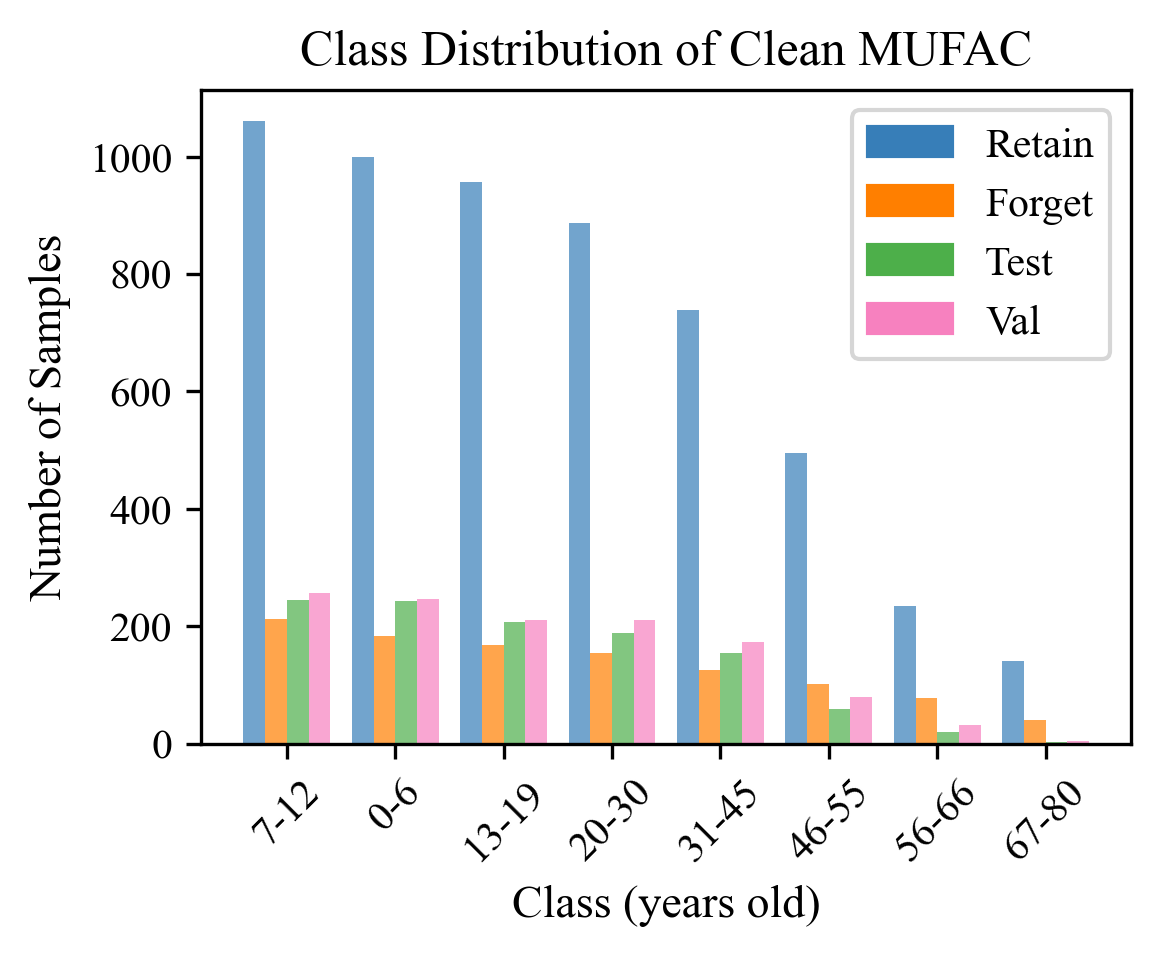}
    \vspace{-1em} 
\caption{\textbf{Number of MUFAC Samples per Class \& Split.} Unlike the balanced CIFAR-10/100 splits, MUFAC exhibits imbalanced class distributions of that varies across the retain, forget, test, and validation splits.\looseness=-1}
\label{fig:mufac_classes}
\end{figure}

Moreover, \Cref{fig:mufac_classes} presents the class distribution of samples in the clean version of MUFAC, showing that the forget set and the unseen set (\ie, the validation set in our case) follow different class distributions. The strong performance of LoTUS in MUFAC further suggest that the assumption of distributional similarity between the forget and unseen sets, discussed in \cref{sec:upper_bound}, can be relaxed.
\looseness=-1

\section{Failure Analysis}\label{sec:failure}
\begin{figure}
    \centering
    \includegraphics[width=0.85\columnwidth]{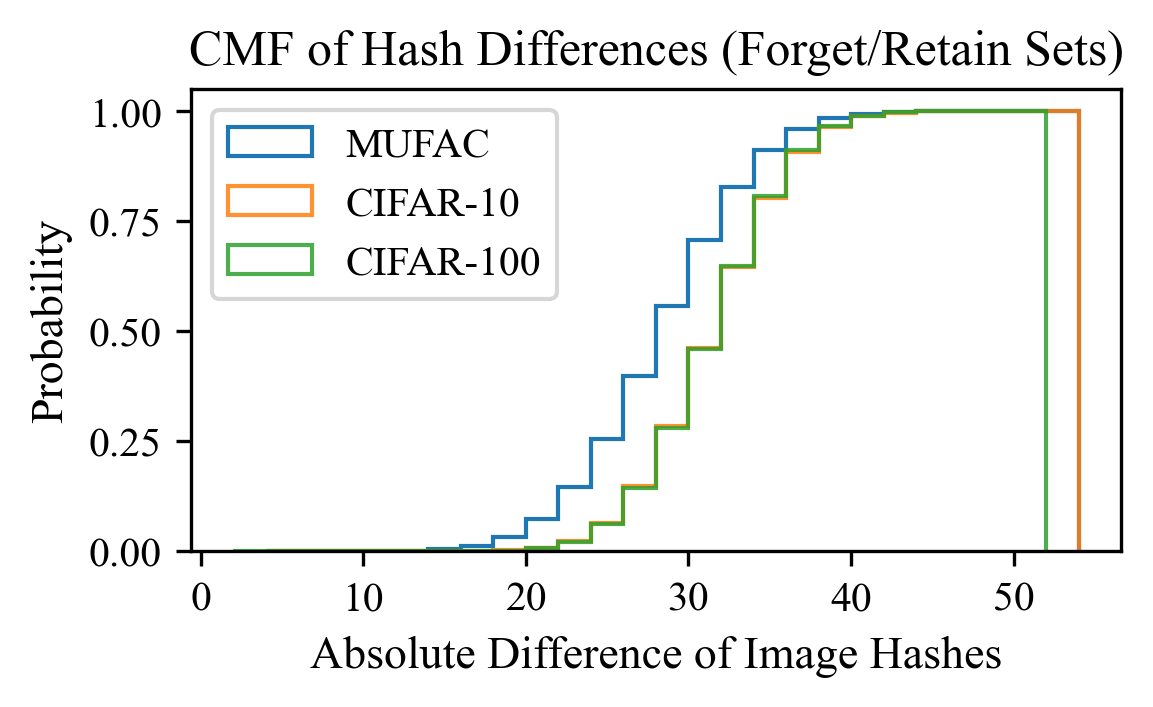}
    \vspace{-1em}
    \caption{\textbf{Orthogonality of Forget/Retain Sets.} We measure the similarity between samples in the forget and retain sets using the absolute difference between their image hashes. MUFAC exhibits significantly higher similarity between forget and retain sets, complicating the unlearning process.\looseness=-1}
    \label{fig:hash_diff}
\end{figure}

Unlearning samples from MUFAC (the clean version) presents greater challenges for all unlearning methods, as reflected in significantly higher JSD scores in \cref{tab:avg_gap}. In addition, MUFAC \& ResNet18 is the only benchmark where LoTUS achieves the second-best Avg Gap rather than the best. To explore the particularities of this dataset, we investigated the orthogonality of the forget and retain sets (\ie, how much they differ). \Cref{fig:hash_diff} presents that the images in the forget and retain sets of MUFAC are more similar, making unlearning more challenging.
\looseness=-1

\section{Class Unlearning with LoTUS}\label{sec:class}
\begin{table*}
\small
\centering
\begin{adjustbox}{width=\textwidth}
\begin{tabular}{l@{\hspace{1ex}}l@{\hspace{1ex}}|ccccccccc>{\columncolor{blue!10}}{c}}
 & Metric ($\downarrow$)
 & Gold Std
 & Finetuning
 & NegGrad+
 & RndLbl
 & BadT
 & SCRUB
 & SSD
 & UNSIR
 & SalUn
 & \tb{LoTUS} \\\toprule
 \multirow{3}{*}{\rotatebox[origin=c]{90}{\makecell{TinyIN \\ Pizza}}}
    & Avg Gap & 0.0000 & 0.2975 & 0.3250 & \ul{0.2925} & 0.3125 & 0.4200 & 0.3650 & 0.5075
    & \ul{0.2925}
    & \tb{0.0925}\\
    & JSD $\!\times\!1e4$ & 0.00\std{0.00} & 94.96\std{7.24} & 86.36\std{9.66} & 92.27\std{6.43} & 72.62\std{22.07} & 73.10\std{0.82} & \ul{34.96\std{14.21}} &  102.29\std{9.33} 
    & 91.01\std{8.59}
    & \tb{37.02\std{18.68}}\\
    & Time (min.)
    & 42.15\std{16.05}
    & 3.23\std{0.01}
    & 3.24\std{0.03}
    & 3.27\std{0.03}
    & 1.59\std{0.01}
    & 4.05\std{0.03}
    & 3.19\std{0.03}
    & \tb{1.01\std{0.01}}
    & 3.98\std{0.01}
    & \ul{1.30\std{0.02 }}\\
    \midrule
\multirow{3}{*}{\rotatebox[origin=c]{90}{\makecell{C-100 \\ Beaver}}}
    & Avg Gap
    & 0.0000 & \ul{0.2825} & 0.3725 & 0.2925 & 0.3000  & 0.3225 & 0.4325 & 0.4050 & \ul{0.2850} & \tb{0.1200}
    \\
    & JSD $\!\times\!1e4$ & 0.00\std{0.00}
    & 101.48\std{2.87} & 108.50\std{2.59} & 102.66\std{3.11} & 78.65\std{3.12} & 64.09\std{8.71} & \ul{45.19\std{9.19}} & 76.28\std{6.88} & 100.93\std{2.44} 
    & \tb{25.46\std{1.41}}\\
    & Time (min.)
    & 4.00\std{0.11}
    & 0.43\std{0.00}
    & 0.44\std{0.01}
    & 0.45\std{0.00}
    & 0.26\std{0.01}
    & 0.55\std{0.00}
    & 0.83\std{0.03}
    & \tb{0.20\std{0.01}}
    & 1.16\std{0.01}
    & \ul{0.23\std{0.01}}\\
    \bottomrule
\end{tabular}
\end{adjustbox}
\caption{\tb{Class Unlearning} with ResNet18 models and the TinyImageNet (TinyIN) and CIFAR-100 (C-100) datasets. We highlight the \tb{best} and \ul{second-best} scores.\looseness=-1}
\label{tab:class}
\end{table*}
\begin{figure}
    \centering
    \begin{subfigure}[b]{0.25\columnwidth}
        \centering
        \includegraphics[width=\textwidth]{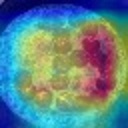}
        \caption{\makecell{prediction:\\ \textit{pizza}}}
        \label{fig:subfig1}
    \end{subfigure}
    \hspace{5em}
    \begin{subfigure}[b]{0.25\columnwidth}
        \centering
        \includegraphics[width=\textwidth]{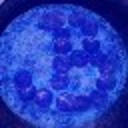} 
        \caption{\makecell{prediction: \\ \textit{potpie}}}
        \label{fig:subfig2}
    \end{subfigure}
    \caption{\tb{Class Activation Maps and Model Predictions:} (a) before and (b) after class unlearning.\looseness=-1}
    \label{fig:gradcams}
\end{figure}
After retraining the model excluding a single \textit{pizza} image from the
training set, the model preserves \emph{global information}
that stems from the remaining \textit{pizzas} in the training set,
being able to correctly classify many of them (see Forget Acc. in \cref{tab:avg_gap_supp}).
In instance-wise unleanring,
LoTUS prevents performance degradation by preventing the elimination of \emph{global information}. To do so, it uses accuracy on labeled unseen \textit{pizzas}, $\text{Acc}(f_\text{orig}, D_u)$ in \cref{eq:tau_d} as an estimator of \emph{global information}.
\looseness=-1

Framing class unlearning as sequential instance-wise unlearning applied to all class samples, \emph{global information} is ultimately eliminated (see Class Activation Maps of \textit{pizza} class in \cref{fig:gradcams}).
Since there is no global information to estimate, we also do not need the unseen set. To adapt LoTUS to class unlearning, we set as objective the accuracy on the forget set to become zero (an empirical observation by retaining the model without the specific class):
\begin{equation}
\tau_d\!=\!\exp\big(\alpha(\text{Acc}(f_\text{un},D_f)\!-\! \cancelto{0}{\text{Acc}(f_\text{orig},D_u))}\big)
\end{equation}

\Cref{tab:class} shows that LoTUS can be adapted to the class unlearning task, outperforming state-of-the-art methods, combining unlearning effectiveness and efficiency.
\looseness=-1

\section{Contribution of Gumbel noise}\label{sec:gumbel}
\begin{table*}
\small
\centering
\begin{adjustbox}{width=\textwidth}
\begin{tabular}{@{}l@{\hspace{1ex}}c|cccc|cccc}
 && \multicolumn{4}{c|}{Vision Transformer} & \multicolumn{4}{c}{ResNet18} \\
 && TinyImageNet & CIFAR-100 & CIFAR-10 & MUFAC & TinyImageNet & CIFAR-100 & CIFAR-10 & MUFAC\\\toprule
 \multirow{2}{*}{\rotatebox[origin=c]{90}{Avg Gap}}
 & \makecell{Gumbel-\\Softmax} & \tb{0.0150} & \tb{0.125} & \tb{0.0050} & \tb{0.0200} & \tb{0.1675} & 0.1200 & \tb{0.0350} & 0.1250 \\\cmidrule{2-10}
 & \makecell{Softmax with \\ Temperature} & 0.0675 & 0.0225 & \tb{0.0050} & \tb{0.0200} & 0.1850 & \tb{0.1075} & 0.0675 & \tb{0.1175} \\\midrule
 \multirow{2}{*}{\rotatebox[origin=c]{90}{JSD$\times1e4$}}
 & \makecell{Gumbel-\\Softmax} & \tb{0.03} & \tb{0.04}  & \tb{0.01} & \tb{0.05} & \tb{0.62} & 1.67 & \tb{0.32} & \tb{6.90} \\\cmidrule{2-10}
 & \makecell{Softmax with \\ Temperature} & 0.15 & \tb{0.04} & \tb{0.01} & 0.08 & 0.65 & \tb{1.36} & 0.41 & 7.33 \\\bottomrule
\end{tabular}
\end{adjustbox}
\caption{\tb{Contribution of Gumbel noise into the activation function.} Ablation analysis using Gumbel-Softmax and Softmax with Temperature as activation functions. LoTUS performs better with Gumbel-Softmax in the majority of the benchmarks.\looseness=-1}
\label{tab:gumbel}
\end{table*}
In \cref{tab:gumbel}, we demonstrate the contribution of the introduction of Gumbel noise in the Softmax activation function. To do so, we perform an ablation analysis using the Gumbel Softmax and the Softmax with Temperature as activation functions in LoTUS. Softmax with Temperature is defined similarly with \cref{eq:probabilities} as:
\begin{equation}
	p_i = s_t(\pi, \tau) = \frac{\exp \left( \left( \log \pi_i \right) / \tau \right)}{ \sum_{j=1}^k \exp \left( \left( \log \pi_j \right) / \tau \right)} ,  \; i = 1,\!\dots\!, k
 \label{eq:probabilities}
\end{equation}

\section{Entropy-based Analysis of the Streisand Effect}\label{sec:entropy_based}
\begin{figure}
    \centering
    \includegraphics[width=0.49\columnwidth]{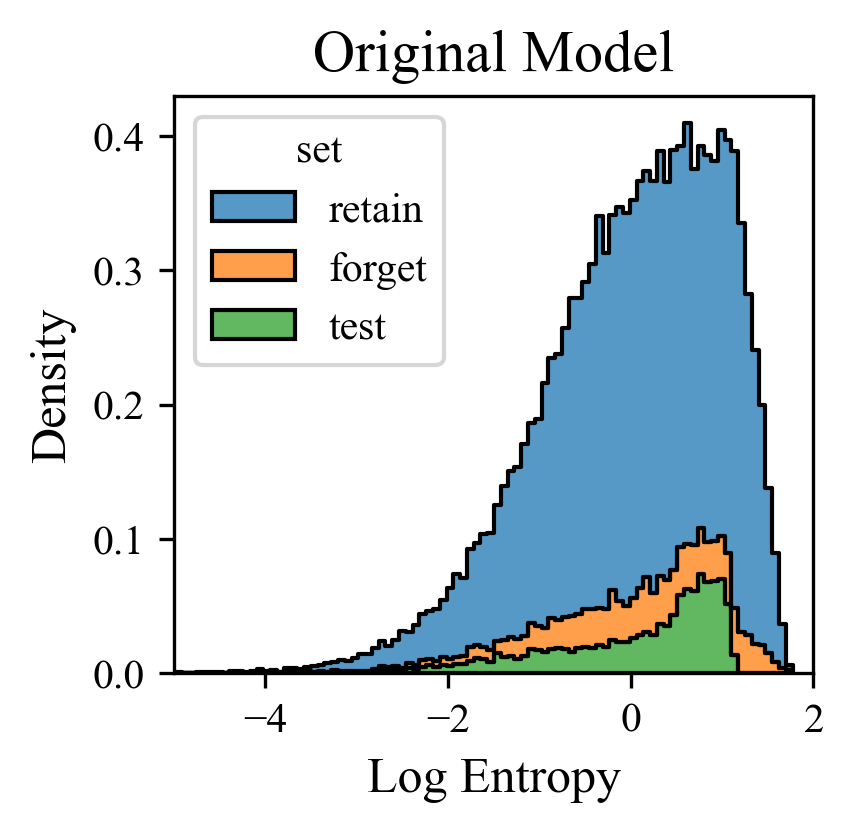}
    \includegraphics[width=0.49\columnwidth]{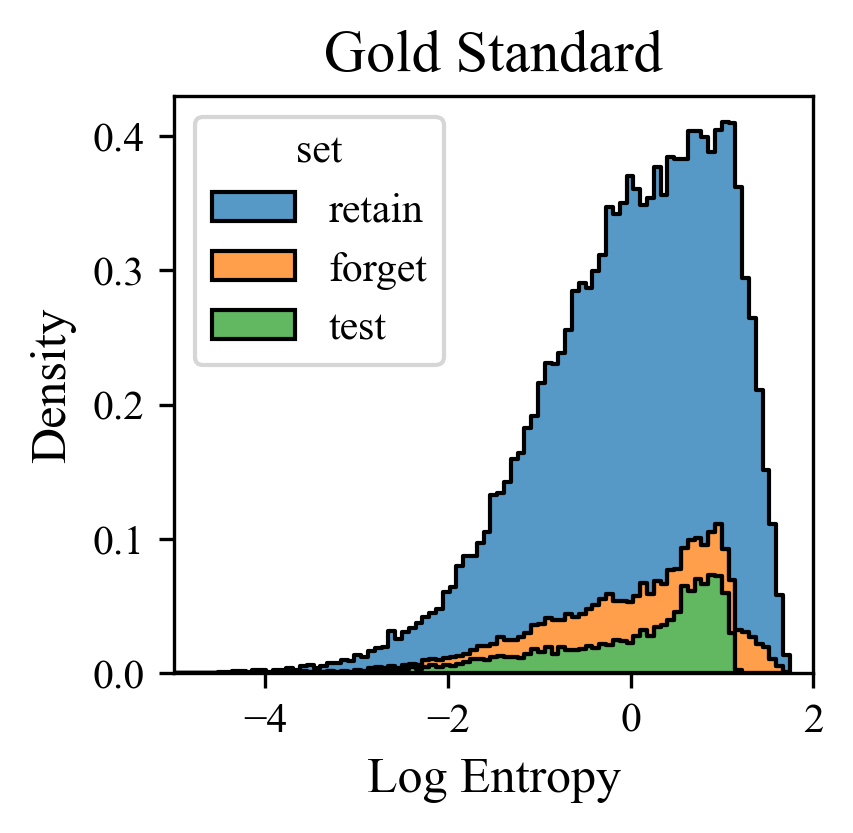}
    \vspace{0.5em}
    \includegraphics[width=0.49\columnwidth]{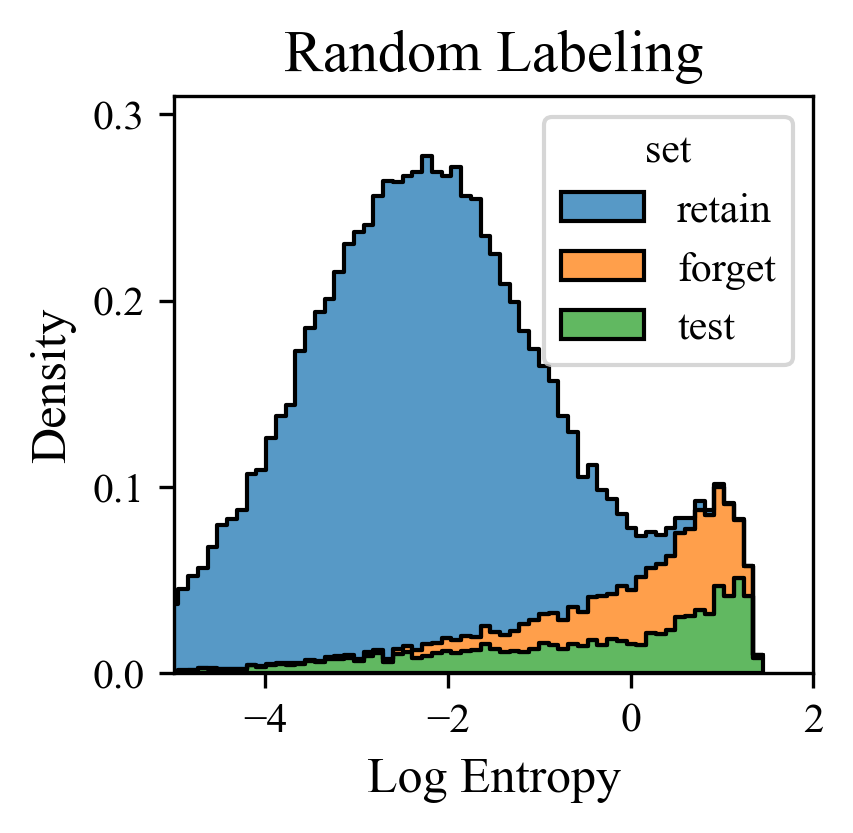}
    \includegraphics[width=0.49\columnwidth]{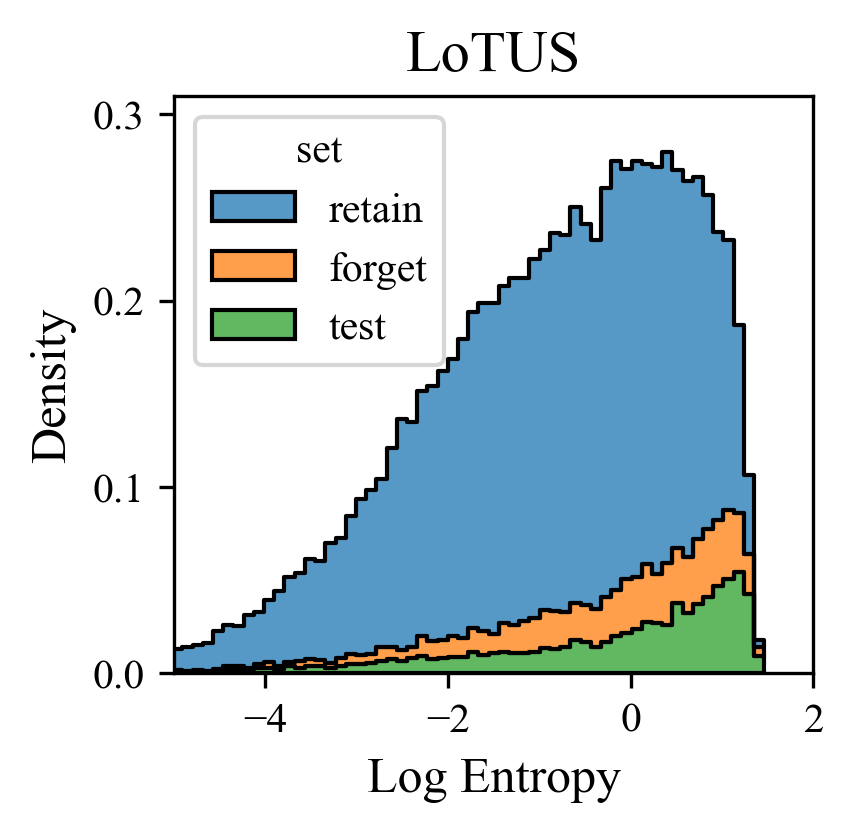}
    \vspace{-1em} 
    \caption{\textbf{Privacy Evaluation via entropy comparison:} LoTUS achieves indistinguishable entropy distributions between forget and retain sets, similar to the orignal and gold standard models. In contrast, Random Labeling produces disproportionately lower entropy in the retain set, making it easier for adversaries to distinguish retain from forget and unseen samples.\looseness=-1}
    
    \label{fig:streisand}
\end{figure}

Further evaluation of the Streisand effect includes investigating the model's uncertainty, as in~\cite{golatkar2020eternal}. In~\cref{fig:streisand}, it is shown that LoTUS prevents an adversary from readily inferring whether an instance is a member of the training set, or whether it belongs to the forget or retain set, since the entropy distributions of the forget/retain/test sets are similar. In contrast, the existing unlearning method~\cite{graves2021amnesiac} that also performs in the output space, but indiscriminately increases the entropy, clearly presents a significant vulnerability to the Streisand effect.
\looseness=-1

\section{Social Impact}
LoTUS can address privacy-related concerns, such as opt-out requests, where users request their data to be deleted not only from the databases, but also from the DNN models. From a security perspective, LoTUS can be applied to unlearn training samples modified by adversaries, which may otherwise compromise the model's performance. In such scenarios, where privacy or security issues arise for specific data points and need to be removed, instance-wise unlearning is more consistent with real-world conditions than class unlearning~\cite{cha2024learning}.
\looseness=-1

\end{document}